\documentclass[lettersize,journal]{IEEEtran}
\usepackage{amsmath,amsfonts}
\usepackage{newtxtext,newtxmath}
\usepackage{algorithmic}
\usepackage{algorithm}
\usepackage{array}
\usepackage[caption=false,font=normalsize,labelfont=sf,textfont=sf]{subfig}
\usepackage{textcomp}
\usepackage{stfloats}
\usepackage{url}
\usepackage{verbatim}
\usepackage{graphicx}
\usepackage{cite}
\usepackage{makecell}
\usepackage[hidelinks]{hyperref}
\usepackage{graphicx}
\usepackage[export]{adjustbox} 
\usepackage{flafter}
\usepackage{soul}
\newenvironment{notetopractitioners}{
    \vspace{-0pt}
    \normalfont\small 
    \bf
    \textit{Note to Practitioners---}\ignorespaces
}{\par\vspace{6pt}} 

\begin{document}

\title{An Adaptive Grasping Force Tracking Strategy for Nonlinear and Time-Varying Object Behaviors}

\author{Ziyang Cheng, Xiangyu Tian, Ruomin Sui, Tiemin Li, Yao Jiang,~\IEEEmembership{Member,~IEEE}
\thanks{This work was supported by the National Natural Science Foundation of China under Grants 52375017 and 52175017, and the Joint Fund of Advanced Aerospace Manufacturing Technology Research under Grant U2037202. \textit{(Corresponding author: Yao Jiang.)}} 
\thanks{Ziyang Cheng is with Weiyang College, Tsinghua University, Beijing 100084, China (e-mail: zy-cheng22@mails.tsinghua.edu.cn).}
\thanks{Xiangyu Tian, Ruomin Sui, Tiemin Li, and Yao Jiang are with the Department of Mechanical Engineering, Tsinghua University, Beijing 100084, China (e-mail: tian-xy20@mails.tsinghua.edu.cn; suirm19@mails.tsinghua.edu.cn; litm@mail.tsinghua.edu.cn; jiangyao@mail.tsinghua.edu.cn).}
}%

\markboth{IEEE Transactions on Automation Science and Engineering,~Vol.~XX, No.~XX, Month~YYYY}%
{Cheng \MakeLowercase{\textit{et al.}}: Adaptive Force Tracking for Nonlinear and Time-Varying Objects}

\IEEEpubid{0000--0000/00\$00.00~\copyright~2021 IEEE}

\maketitle

\begin{abstract}
Accurate grasp force control is one of the key skills for ensuring successful and damage-free robotic grasping of objects. Although existing methods have conducted in-depth research on slip detection and grasping force planning, they often overlook the issue of adaptive tracking of the actual force to the target force when handling objects with different material properties. The optimal parameters of a force tracking controller are significantly influenced by the object's stiffness, and many adaptive force tracking algorithms rely on stiffness estimation. However, real-world objects often exhibit viscous, plastic, or other more complex nonlinear time-varying behaviors, and existing studies provide insufficient support for these materials in terms of stiffness definition and estimation. To address this, this paper introduces the concept of generalized stiffness, extending the definition of stiffness to nonlinear time-varying grasp system models, and proposes an online generalized stiffness estimator based on Long Short-Term Memory (LSTM) networks. Based on generalized stiffness, this paper proposes an adaptive parameter adjustment strategy using a PI controller as an example, enabling dynamic force tracking for objects with varying characteristics. Experimental results demonstrate that the proposed method achieves high precision and short probing time, while showing better adaptability to non-ideal objects compared to existing methods. The method effectively solves the problem of grasp force tracking in unknown, nonlinear, and time-varying grasp systems, {demonstrating the generalization capability of our neural network and} enhancing the robotic grasping ability in unstructured environments.
\end{abstract}

\begin{notetopractitioners}
In unstructured scenarios such as daily life and agriculture, robots often need to adaptively control grasping force to safely handle objects with unknown properties like weight or hardness. Based on existing methods for determining the desired grasping force, this work focus on controlling actual grasping force to dynamically track the desired force. In designing this force tracking controller, object stiffness significantly impacts optimal parameters. Many adaptive controllers rely on stiffness estimation, but real-world objects often exhibit nonlinear, time-varying behaviors such as viscosity and plasticity, making existing estimators designed for ideal elastic bodies prone to errors. To address this, we propose a generalized stiffness concept suitable for non-ideal objects and a corresponding LSTM-based online generalized stiffness estimator, enabling practitioners to achieve adaptive parameter tuning for the force tracking controller. As an example, we implement this estimator on a PI controller. Experiments demonstrate our controller's strong adaptability across elastic, viscoelastic, plastic, and variable-stiffness objects, achieving high accuracy and fast tracking speed in grasping tasks.
\end{notetopractitioners}

\begin{IEEEkeywords}
Force tracking, generalized stiffness estimation, grasp force control, LSTM
\end{IEEEkeywords}

\section{Introduction}
\label{sec:chapter1}
\IEEEPARstart{G}{rasping} is one of the most important skills for robots. To ensure successful and non-damaging object grasping, it is necessary to precisely control the grasp force applied to the object \cite{ref1}. In structured environments such as automated production lines, robots are used for repetitive grasping of objects with fixed characteristics, such as weight and shape, and the required grasp force can be pre-calculated. Thus, grasp force control can be achieved through pre-programming or teaching methods \cite{ref2}. {Even for challenging soft robotic manipulators, many studies achieve effective control by modeling their dynamics} \cite{ref22, ref23, ref24}, while external forces and object properties remain relatively constant or inapparent, making them easy to model or neglect. However, with the increasing application of robots in unstructured environments such as daily life and agriculture, where objects vary greatly and are subject to complex environmental constraints (e.g., fruits being pulled by their stems), the unknown object properties and external forces present significant challenges for grasp force control \cite{ref3,ref4}.

\IEEEpubidadjcol
Grasp force control in unstructured environments can be divided into two aspects: decision-making and execution. In decision-making, force planning is required. Since objects may be subjected to varying external forces, the minimum grasp force required to prevent the object from slipping needs to be calculated dynamically and in real-time, based on which the target grasp force is determined. As mentioned in \cite{ref5}, it is crucial to detect slippage after the robot has made contact with the object. For example, Liu \textit{et al.}~\cite{ref6} use Short-Time Fourier Transform and Discrete Wavelet Transform algorithms to process contact force signals and detect slippage using machine learning methods, thereby determining the optimal grasp force for delicate fruits; Su \textit{et al.}~\cite{ref7} detect and classifies slippage through fingertip contact forces and micro-vibrations to plan the required grasp force; Zhang \textit{et al.}~\cite{ref8} propose a method based on the local deformation degree for visual-tactile sensors, which can quantify the initial slip degree on objects with complex contact properties, laying the foundation for force planning; James \textit{et al.}~\cite{ref9} use Support Vector Machines to assess the slippage state and determine the minimum force required for grasping an object; Muthusamy \textit{et al.}~\cite{ref10} introduce a feature-based slippage detection method for neuromorphic vision sensors and adjusts the grasp force accordingly to suppress slippage. Thus, in decision-making, slippage detection and slip-based grasp force planning have been thoroughly researched.

In terms of execution, force tracking is necessary to ensure that the actual force applied to the object follows the target force dynamically in real time. Since both excessive and insufficient grasp force can damage or cause the object to fall, not only must steady-state errors be kept within a tolerable range, but convergence speed should also be maximized to prevent failure due to overshooting or lag in the actual force. However, many studies \cite{ref6,ref7,ref8,ref9,ref11} focus only on slippage detection or force planning, neglecting force tracking methods. Some studies do mention it, such as Zhou \textit{et al.}~\cite{ref12} using a PID controller to regulate actual force, Hakkak \textit{et al.}~\cite{ref13} employing a PI controller, and Muthusam \textit{et al.}~\cite{ref10} using a fuzzy controller. However, these methods lack adaptability and require offline tuning of optimal controller parameters based on the object's actual properties to achieve good convergence speed and tracking accuracy. Since object properties are unknown in unstructured environments, controller parameters cannot be predetermined, making it impossible to achieve optimal force tracking performance using these methods. 

To solve this problem, some research uses sensor data that contains the mechanical characteristics of the object, such as the displacement and force applied to robotic fingers, to estimate system parameters online and adjust the controller accordingly. For example, Treesatayapun~\cite{ref17} proposes a Multi-input Fuzzy Rules Emulated Network (MiFREN) method that can continuously optimize its parameters during grasping. This method can theoretically handle objects with unknown mechanical behaviors, but the network includes 25 parameters that need to be adjusted online, requiring extensive data collection to achieve optimal parameter tuning, leading to longer probing times during initial grasping. Wang \textit{et al.}~\cite{ref26} {propose the Input Convex Long Short-Term Memory Network (ICLSTM), which can predict the next system state given a control attempt, providing valuable information for adaptive control. However, it cannot directly guide the controller’s adjustment. Consequently, a Model Predictive Control (MPC) approach is needed with ICLSTM rollouts at each step, incurring significant overhead and limiting its use to low-frequency applications such as chemical and energy systems.}

Therefore, to reduce probing time, it is necessary to minimize the number of parameters and focus on the object characteristic parameters that have the most significant impact on the controller {and can guide the controller adjustment in a straightforward manner}. Liu \textit{et al.}~\cite{ref27} and Kumar \textit{et al.}~\cite{ref28} {have employed reinforcement learning to train encoders that automatically extract low-dimensional representations of environmental information for direct use by the control policy; however, in our scenario, such a feature—stiffness—can be explicitly defined.} The essence of force tracking is to control the movement of the robotic finger in such a way that the object generates the desired reactive force, and the stiffness of the object reflects the relationship between deformation and applied force. Thus, stiffness is crucial for force tracking controllers. From a biomimetic perspective, this conclusion also holds: humans use multi-sensory coordination to judge the stiffness of an object \cite{ref18}, which has significant implications for human hand grasp control \cite{ref19}. 

Many studies adjust the controller by considering the stiffness of the object, {but numerous challenges and issues still persist.} De Schutter \textit{et al.}~\cite{ref14} indicate that object stiffness can effectively characterize system properties and designed a PI controller that adjusts gain based on stiffness, but did not provide a method for online stiffness estimation. Zhang \textit{et al.}~\cite{ref15} estimate object stiffness based on initial grasp information and adjusts the parameters of a fuzzy logic controller accordingly. Unfortunately, this estimation is only performed during the initial grasp, assuming that the object has fixed stiffness and is a linear, time-invariant system. Once the object stiffness changes (e.g., due to varying grasp forces), the initially estimated stiffness becomes inaccurate, leading to a decline in controller tracking performance. Wei \textit{et al.}~\cite{ref16} propose a variable impedance force controller that defines object stiffness based on an ideal elastic body model and designs an online stiffness estimator. The impedance parameters of the controller are automatically adjusted according to the estimated stiffness. However, for complex objects with nonlinear or time-varying behaviors, the force-displacement relationship does not follow the ideal elastic body model. If force and displacement data are interpreted according to this model, the stiffness will be incorrectly estimated, causing controller parameter mismatch. In summary, while many studies have considered object stiffness and adjusted controller parameters accordingly, their stiffness definitions and estimation methods are not applicable to more complex nonlinear time-varying systems, resulting in suboptimal performance for real-world objects. In this context, Wu \textit{et al.}~\cite{ref25} {inspired us by demonstrating the powerful capability of neural networks to estimate specific object parameters from observations of nonlinear systems. However, unlike the object mass estimated in this work, the stiffness of many objects is highly time-varying, which necessitates a more sophisticated design to enable neural networks to perform real-time and accurate estimation.}

In conclusion, to address the force tracking problem in grasping systems with unknown, nonlinear, and time-varying behaviors, we define the generalized stiffness as the key parameter upon which the controller's adaptive parameter adjustment relies. We design an online stiffness estimator and implement a force tracking controller that automatically adjusts parameters based on generalized stiffness. Given the widespread application of two-fingered planar robotic hands in the field of robotics, we take their grasp force tracking tasks as the subject of our study. Chapter \ref{sec:chapter2} proposes a nonlinear time-varying grasping force tracking system model and defines generalized stiffness based on this model. Chapter \ref{sec:chapter3} designs an accurate and fast-converging online generalized stiffness estimator based on LSTM, elaborating on the sources of training data and presenting network training and validation. In Chapter \ref{sec:chapter4}, under the principle of generality, a common PI controller is selected to implement a controller with automatic parameter tuning based on generalized stiffness. We analyze the conditions for the system's stability and theoretically verify the significance of generalized stiffness for the controller. Chapter \ref{sec:chapter5} demonstrates through comparative performance experiments and grasping application tests that the definition and estimation of generalized stiffness enable the controller to adjust parameters online, ensuring both high precision and short probing time while exhibiting superior adaptability.

\section{Grasping Force Tracking System Model}
\label{sec:chapter2}
\begin{figure}[t]
    \centering
    \includegraphics[width=8.8cm]{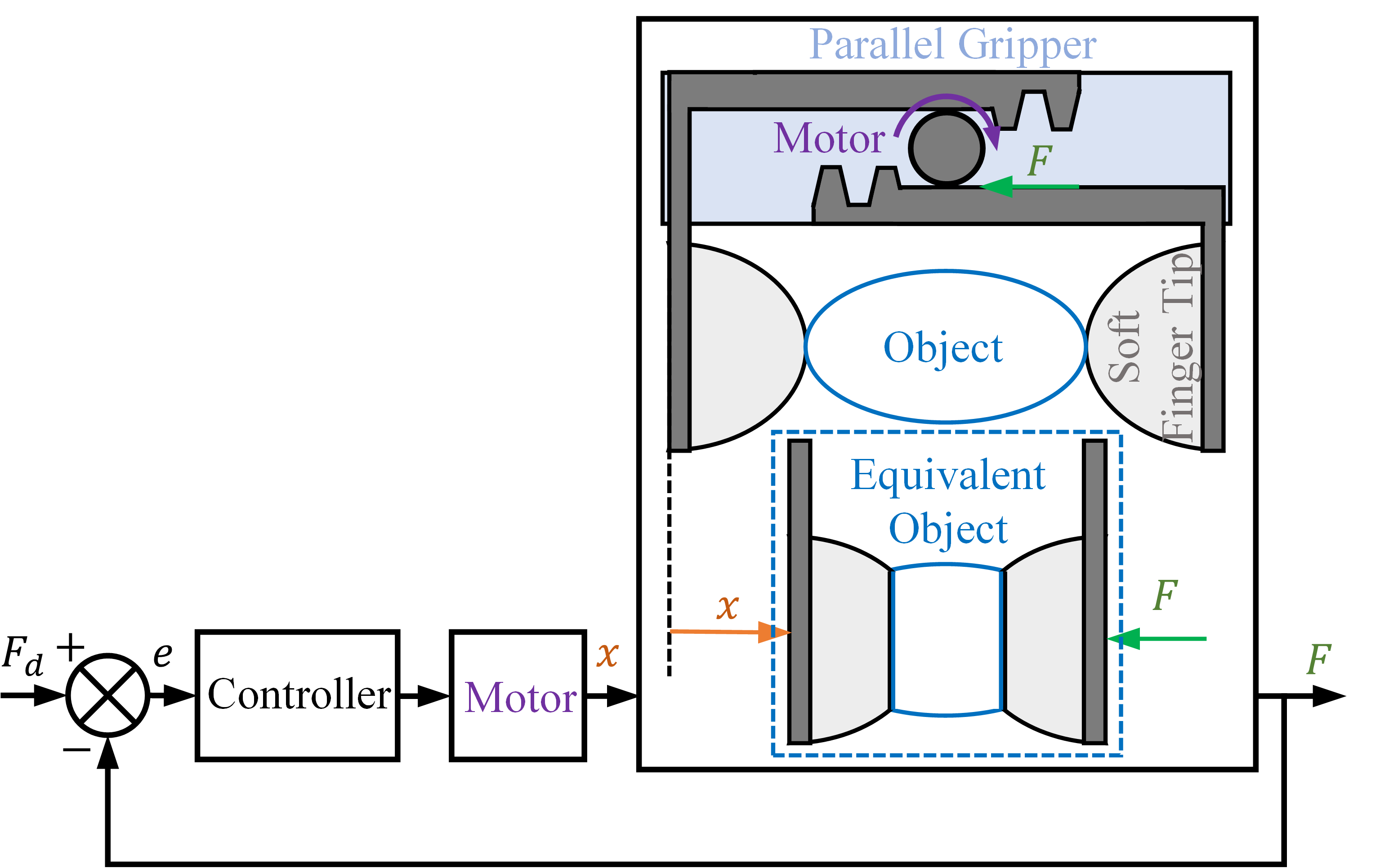}
    \caption{Grasping force tracking system.}
    \label{fig:force_tracking_system}
\end{figure}

In autonomous robotic grasping tasks, force tracking is required to control the actual grasping force. Fig.~\ref{fig:force_tracking_system} shows the control block diagram of the force tracking system for a single-degree-of-freedom two-finger planar gripper. The essence of the system is to control the movement of the gripper fingers to adjust the contact force between the fingers and the object. The object being grasped and the fingertips of the gripper are connected in series, forming an equivalent object, which is treated as the controlled object in the force tracking control system. Therefore, the force-deformation relationship model of this equivalent object (referred to as the object model) plays a crucial role in determining the controller parameters.

Let \( t \) be time, and \( T \) be the discrete control time interval. At \( t = 0 \), the gripper makes contact with the object. The force applied to the object in the direction of the gripper's translational degree of freedom is denoted as \( F(t) \), and the desired target force is \( F_d(t) \). The displacement of the gripper from contact to time \( t \) is \( x(t) \). First, consider an ideal case where the object has variable stiffness, and the displacement due to deformation is entirely determined by the force currently applied to the object:
\begin{equation}
x(t) = f(F(t)) \label{eq1 ideal}
\end{equation}

At any given moment, the stiffness \( k_E(F) = 1/\frac{\mathrm{d}f(F)}{\mathrm{d}F} \) reflects the relationship between the input (displacement) and output (force) of the controlled object, and thus determines the controller parameters. Assume that at \( t = 0 \), the contact occurs and the displacement zero point is calibrated, i.e., \( f(0) = 0 \), so \eqref{eq1 ideal} can be rewritten as:
\begin{equation}
x(t) = \int_0^{F(t)} \frac{\mathrm{d}F}{k_E(F)} \label{eq2 ideal int}
\end{equation}

However, for non-ideal objects that are nonlinear and time-varying, the deformation is not solely determined by the current grasping force, but also depends on time and the forces applied at past times. Therefore, for such objects, the relationship between displacement and force in the object model can be expressed as follows:
\begin{equation}
x(t) = f(t, \mu(t), F(t)) \label{eq3 nonideal}
\end{equation}
where \( f \) is the unknown system response characteristic, and \( \mu(t) = [F(t-T), F(t-2T), \dots, F(0)] \) represents the forces applied to the object at previous time steps. 

In analogy to the ideal variable-stiffness object, we define the generalized stiffness \( k(t, \mu(t), F) = 1/\frac{\partial f(t, \mu(t), F)}{\partial F} \). Compared to traditional stiffness, the generalized stiffness can be defined for nonlinear time-varying systems. It not only depends on the current applied force but may also vary over time and with the forces applied at past moments. Nevertheless, it shares similar physical significance with traditional stiffness and can reflect the relationship between input and output of the controlled object, thus playing a decisive role in determining the controller parameters.

Taking the partial derivative of the right-hand side of \eqref{eq3 nonideal} with respect to \( F(t) \) and integrating, we obtain:
\begin{equation}
x(t) = \int_0^{F(t)} \frac{\mathrm{d}F}{k(t, \mu(t), F)} + C(t, \mu(t)) \label{eq4 nonideal int}
\end{equation}

Since at \( t = 0 \), the gripper makes contact with the object, we have \( F(0) = 0 \), \( x(0) = 0 \), and \( C(0, \mu(0)) = 0 \).

\begin{figure}[t]
    \centering
    \includegraphics[width=8.8cm]{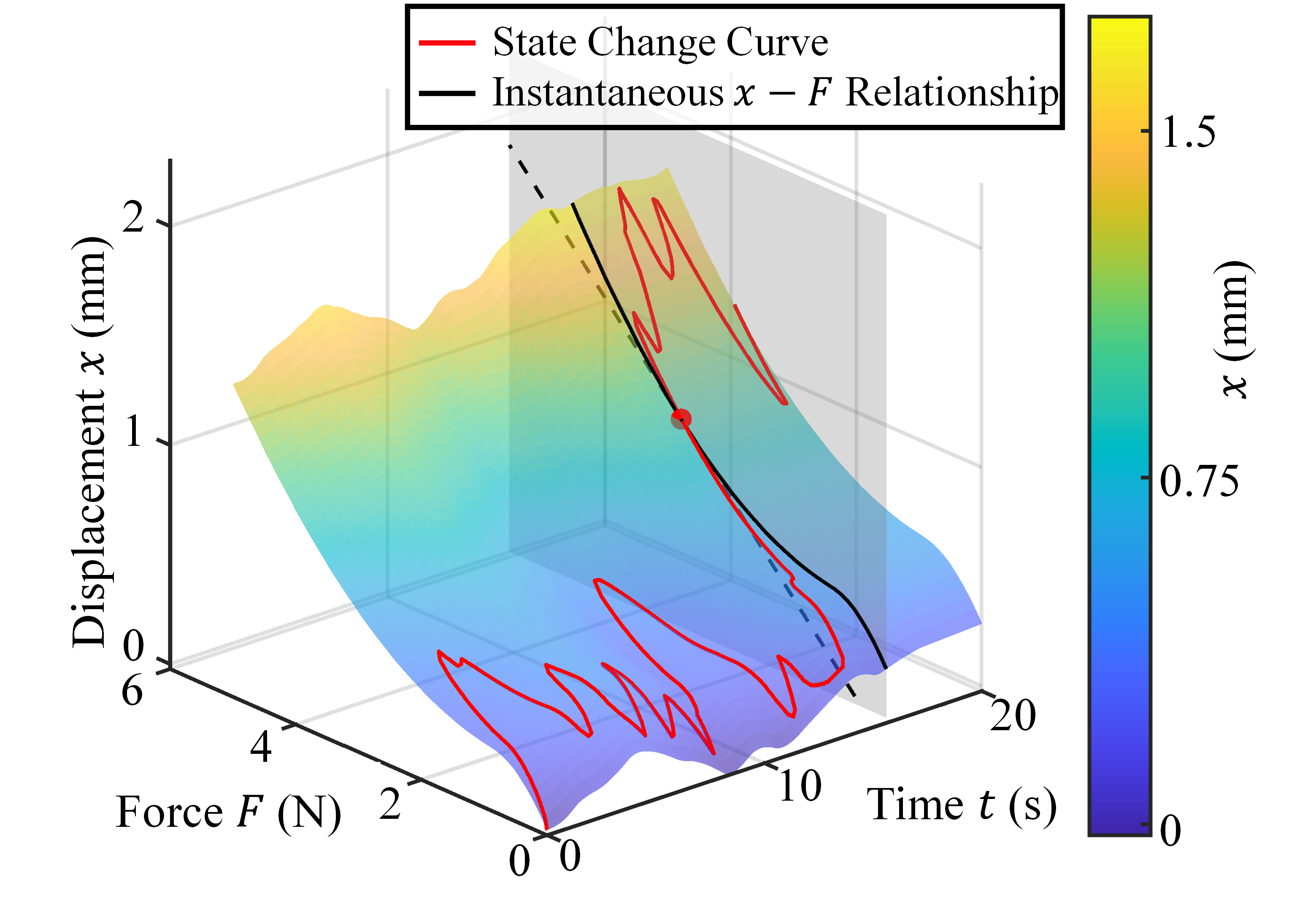}
    \caption{{State change of the object.}}
    \label{fig:state_change_diagram}
\end{figure}

Based on the model proposed, a state diagram of an object is shown in Fig.~\ref{fig:state_change_diagram}. For any instantaneous time \( t \), a 2D curve can be plotted to represent the variation of \( x(t) \) with respect to \( F(t) \). The inverse of the slope of this curve at the current grasping force \( F(t) \) represents the generalized stiffness. As time progresses, a series of such 2D curves form a surface in 3D space. In this context, the actual grasping force \( F(t) \) follows a curve in this surface, and the resulting red curve in the diagram represents the state changes actually experienced by the object.

Thus far, \eqref{eq4 nonideal int} applies to any general continuous nonlinear time-varying system. For the sake of convenience in subsequent analysis, we now make certain assumptions about the object model based on the behaviors of actual objects.

\begin{enumerate}
    \item\textbf{High Control Frequency}
    \label{assumption1}
    
As \( T \to 0 \), the discrete system approaches a continuous system. Taking the total derivative of \eqref{eq4 nonideal int} with respect to \( x \):
\begin{align}
\mathrm{d}x &= \frac{1}{k(t, \mu(t), F(t))} \, \mathrm{d}F(t) \notag \\
&+ \int_0^{F(t)} \frac{\partial \left( \frac{1}{k(t, \mu(t), F)} \right)}{\partial t} \, \mathrm{d}t \, \mathrm{d}F + \mathrm{d}C(t, \mu(t)) 
\label{eq6 High Control Frequency}
\end{align}

Therefore, in the case of a high control frequency, the discrete system can be approximated as follows:
\begin{align}
&x(t+T) - x(t)  \notag\\
=&\frac{1}{k(t, \mu(t), F(t))} \, (F(t+T) - F(t)) \notag \\
+& \int_0^{F(t)} \left( \frac{1}{k(t+T, \mu(t+T), F)} - \frac{1}{k(t, \mu(t), F)} \right) \, \mathrm{d}F \notag \\
+& \left( C(t+T, \mu(t+T)) - C(t, \mu(t)) \right)
\label{eq7 High Control Frequency}
\end{align}

    \item\textbf{Positive Bounded Generalized Stiffness}
    \label{assumption2}
    
The generalized stiffness \( k(t, \mu(t), F) = 1/\frac{\partial f_p(t, \mu(t), F)}{\partial F} \) reflects the partial derivative of displacement with respect to the applied force. For typical objects in real-world scenarios, an increase in the applied force results in an increase in displacement in the direction of the applied force. Therefore, the generalized stiffness is positive.

Since the equivalent object consists of the grasped object and the gripper's elastic fingertips in series, it can be shown that the generalized stiffness of the equivalent object does not exceed the generalized stiffness of the fingertips. Therefore, even for objects like steel, which undergo minimal deformation, we can still ensure that the upper bound of the generalized stiffness is the stiffness of the fingertips. Thus, we have:
\begin{equation}
0 < k(t, \mu(t), F) < k_m \label{eq5 Generalized Stiffness is Positive and Bounded}
\end{equation}

\item\textbf{Positive Continuous Bounded Grasping Force} 
\label{assumption3}

In the anticipated actual grasping process, both the desired and the actual grasping force is positive and does not exceed the maximum force limit of the gripper. Furthermore, the rate of change of force in physical processes is limited and typically does not undergo abrupt changes. Thus, we have:
\begin{equation}
0 < F(t),F_d(t) < F_m \label{eq8 Positive Continuous Bounded Grasping Force}
\end{equation}
\begin{equation}
F(t+T) - F(t) = \epsilon_F(t), \quad |\epsilon_F(t)| < \delta_F \label{eq9 Positive Continuous Bounded Grasping Force}
\end{equation}
\begin{equation}
F_d(t+T) - F_d(t) = \epsilon_{F_d}(t), \quad |\epsilon_{F_d}(t)| < \delta_{F_d} \label{eq9 Positive Continuous Bounded desired Force}
\end{equation}

\item\textbf{Slow and Bounded Zero-Force Drift} 
\label{assumption4}

From \eqref{eq4 nonideal int}, when \( F(t) = 0 \), we have \( x(t) = C(t, \mu(t)) \). This means that even when the force applied to the object approaches zero, the object will still undergo deformation \( C(t, \mu(t)) \). This phenomenon is referred to as zero-force drift. Although some objects exhibit this behavior (such as the chest of an animal undergoing breathing movements under no external force), the drift is slow and bounded:
\begin{align}
C(t+T, \mu(t+T)) - C(t, \mu(t)) &= \epsilon_C(t), \notag\\|\epsilon_C(t)| &< \delta_C \label{eq10 Zero Force Drift is Slow and Bounded}
\end{align}
\begin{equation}
|C(t, \mu(t))| < \Delta_C \label{eq11 Zero Force Drift is Slow and Bounded}
\end{equation}

\item\textbf{Slow Constant-Force Drift} 
\label{assumption5}

According to \eqref{eq7 High Control Frequency}, even if the force remains constant from one time step to the next, \( F(t+T) = F(t) \), the deformation of the object will still change. This change is given by:
\begin{align}
&x(t+T) - x(t) \notag\\
=&\int_0^{F(t)} \left( \frac{1}{k(t+T, \mu(t+T), F)} - \frac{1}{k(t, \mu(t), F)} \right) \, \mathrm{d}F \notag \\
+& \left( C(t+T, \mu(t+T)) - C(t, \mu(t)) \right)
\label{eq12 Constant Force Drift is Slow}
\end{align}

This change is referred to as constant-force drift, which is caused by the variations in generalized stiffness and zero-force drift. Although some objects (such as viscoelastic materials) exhibit this behavior, we assume that the drift is slow, i.e., the constant-force drift occurring within the very short control period \( T \) is a bounded small quantity. In \eqref{eq12 Constant Force Drift is Slow}, \( C(t+T, \mu(t)) - C(t, \mu(t)) = \epsilon_C(t) \) is a bounded small quantity, so the other term is also bounded:
\begin{align}
&\int_0^{F(t)} \left( \frac{1}{k(t+T, \mu(t+T), F)} - \frac{1}{k(t, \mu(t), F)} \right) \, \mathrm{d}F \notag\\
=& \epsilon_{kF}(t), \quad |\epsilon_{kF}(t)| < \delta_{kF}
\label{eq13 Constant Force Drift is Slow}
\end{align}

\item\textbf{Slow and Bounded Generalized Stiffness Drift} 
\label{assumption6}

For the grasping system, although system characteristics may drift, the changes are generally slow and limited. When the generalized stiffness \( k \) is viewed as a function of \( F \), it changes slowly and is bounded as time progresses and the history information \( \mu(t) \) evolves. Thus, we have:
\begin{align}
\frac{k(t+T, \mu(t+T), F) - k(t, \mu(t), F)}{k(t, \mu(t), F)} &= \epsilon_k(t, F), \notag \\
|\epsilon_k(t, F)| &< \delta_k < 1
\label{eq14 Slow and Bounded Generalized Stiffness Drift}
\end{align}
\begin{equation}
\frac{|k(t, \mu(t), F) - k(0, \mu(0), F)|}{k(0, \mu(0), F)} < \Delta_k \label{eq15 Slow and Bounded Generalized Stiffness Drift}
\end{equation}

\end{enumerate}

The above assumptions describe a nonlinear time-varying model that can be used to describe the controlled object in the vast majority of grasping force tracking systems. Based on this model, we define the generalized stiffness and clarify how it determines the controller parameters over short time periods. In the next section, we will design a strategy to estimate the generalized stiffness online, robustly, and accurately for grasping systems conform to this model.

\section{Generalized Stiffness Estimation}
\label{sec:chapter3}
According to the analysis in Chapter \ref{sec:chapter2}, generalized stiffness can characterize the controlled object over an instantaneous period, thereby determining the optimal parameters for the controller. Therefore, in order to achieve online automatic optimization of the controller parameters and realize a high-precision, fast-convergence force tracking controller, it is necessary to achieve accurate online estimation of generalized stiffness within a short probing time. This section analyzes the issues faced by traditional stiffness estimation methods when estimating generalized stiffness and then proposes a generalized stiffness estimation method based on LSTM neural network, followed by training and validation of the network.

\subsection{Design of Generalized Stiffness Estimator}
During the grasping process, the grasping force \( F(t) \) and the position of the robotic finger \( x(t) \) are measurable data. Due to the complex behaviors of the object, traditional linear fitting methods are no longer applicable. Thus, neural networks are considered for stiffness estimation. Additionally, based on the idea of residual connections, a rough stiffness estimation method is designed as a physical heuristic for the neural network, which can reduce the network size and accelerate training.

For an ideal variable-stiffness object, the variation of \( F(t) \) and \( x(t) \) satisfies \eqref{eq2 ideal int}. A commonly used stiffness estimation method is to fit the slope of the latest \( n \) data points on the \( F \)-\( x \) data plot using the least squares method \cite{ref15,ref16}, with the estimated slope approximating the stiffness at the current moment:
\begin{equation}
\hat{k}_E(t) = \text{slope} \approx k_E(F(t))
\end{equation}

However, when estimating generalized stiffness, we aim to estimate \( k(t, \mu(t), F(t)) \) rather than \( k_E(t) \). As shown in \eqref{eq7 High Control Frequency}, when the force is applied slowly, i.e., \( F(t) - F(t-T) \) is small, zero-force drift and constant-force drift dominate the variation of \( x(t) \), making the slope of the \( F \)-\( x \) data plot an inaccurate estimate of stiffness.

Thus, we make a correction: when the force loading rate is below a threshold, the estimated value is not updated, but instead the historical value is retained:
\begin{equation}
\hat{k}_E(t) = \hat{k}_E(t-T), \quad \text{if} \quad F(t) - F(t-T) < \text{threshold} 
\end{equation}

This gives the stiffness estimation method based on the Least Squares Method (this improved method is referred to as LSM hereafter). Although we have made some corrections, when the force loading rate remains slow for a long time, the stiffness estimation cannot be updated, leading to an inability to perceive changes in the stiffness of the variable-stiffness object. In conclusion, this method has limitations when the force loading rate is slow and zero-force drift and constant-force drift are significant, but it offers reasonable accuracy in other cases and can serve as a physical heuristic for the neural network.

\begin{figure}[t]
\centering
\includegraphics[width=8.8cm]{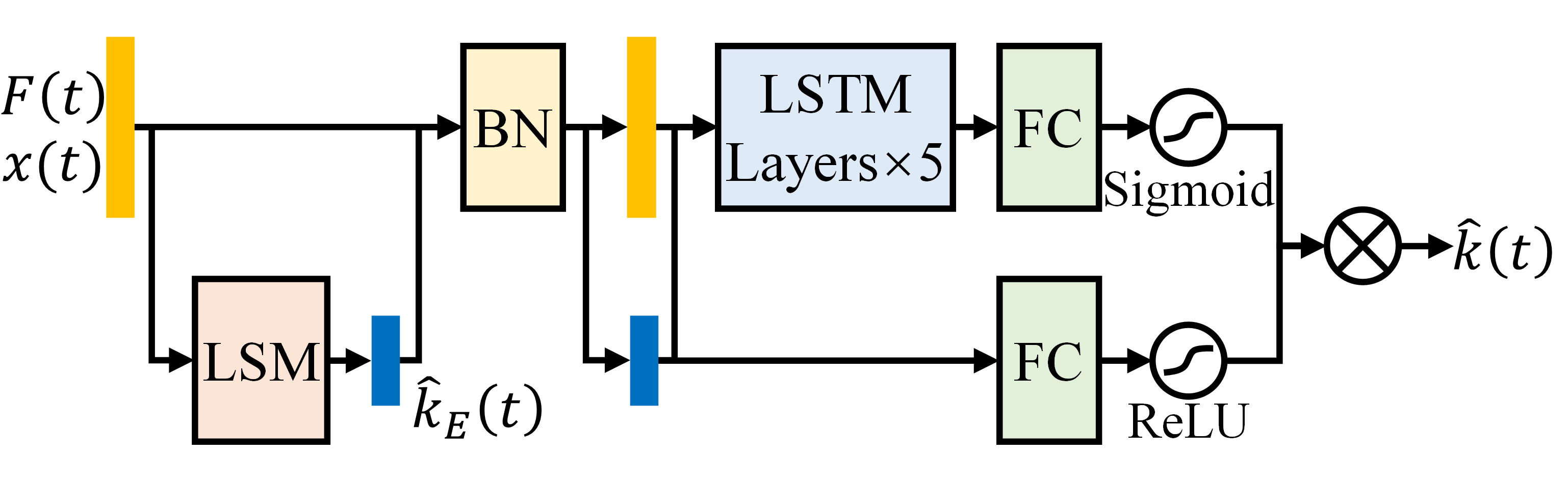}
\caption{Neural network workflow. {$F$ is the grasping force, $x$ is the finger displacement, $\hat{k}_E$ is the stiffness estimated by LSM, and $\hat{k}$ is the generalized stiffness estimated by our method.}}
\label{fig:LSTM}
\end{figure}

Since both the input data and output estimates are time series, which align with the input-output characteristics of an LSTM network, we employ LSTM to fit the residuals based on the physical heuristic of the Least Squares Method (LSM). The output of the combined network is represented as \( \hat{k} \). The specific design is shown in Fig.~\ref{fig:LSTM}.

The LSTM consists of five layers, each with a hidden dimension of 512. {Additionally, the network includes two Fully Connected (FC) layers. The first FC layer has an input dimension of 512 and, at each time step, receives the hidden state from the final LSTM cell, producing a one-dimensional vector as its output. The second FC layer, with both input and output dimensions of 1, serves to scale the output of the LSM.} To correct the least squares stiffness estimator, the core idea is to use the output of the LSTM (and the FC layer behind) as a correction factor, which is multiplied by the output of the physical heuristic. To eliminate the effect of Batch Normalization (BN) and to avoid the issue where the correction factor processed by the Sigmoid function is less than 1, leading to \( \hat{k}(t) \) always being smaller than \( \hat{k}_E(t) \), we apply a FC layer to \( \hat{k}_E(t) \) and use ReLU as the activation function to ensure that \( \hat{k}(t) > 0 \) after multiplication.

In the design of the loss function, we are concerned with the ratio \( \eta(t) = \hat{k}(t)/k(t, \mu(t), F(t)) \) between the generalized stiffness estimate and the true value, and we desire the loss to be zero when \( \eta(t) = \text{1} \), and for the loss to tend to a maximum when \( \eta(t) \) approaches zero or infinity. Conventional loss functions like Mean Squared Error (MSE) do not meet these requirements, so we designed the following loss function:
\begin{equation}
\text{loss} = \text{Mean}_t \left( \eta(t)^2 +  \frac{1}{\eta(t)^2} - 2 \right) 
\end{equation}

\subsection{Data Acquisition, Training and Validation}
To train the neural network, we need to acquire input data \( [F(t), x(t)] \) and corresponding target outputs \( k(t, \mu(t), F(t)) \). Since the adaptive grasping scenario in this paper requires the generalized stiffness estimator to work for various objects, we hope the estimator performs well for all systems that meet the assumptions in Chapter \ref{sec:chapter2}, where we defined a comprehensive system model that nearly encompasses the wide array of nonlinear time-varying grasping processes observed in practice, using Equation~\eqref{eq4 nonideal int} {along with specific constraints on the physical quantities. Consequently, almost all real-world grasping force tracking processes—including variations in force, displacement, and the object’s generalized stiffness—fall within this model.}

{Therefore, we generate the training data by randomly sampling multiple grasping processes from this system model using an empirical distribution. Any real-world force-tracking process that adheres to our system model will benefit from interpolation generalization—even if its distribution deviates slightly from the empirical one. This interpolation generalization is relatively straightforward for neural networks,} and is also tested in Chapter \ref{sec:chapter5}.

\begin{figure*}[htbp]
\centering
\includegraphics[width=17cm]{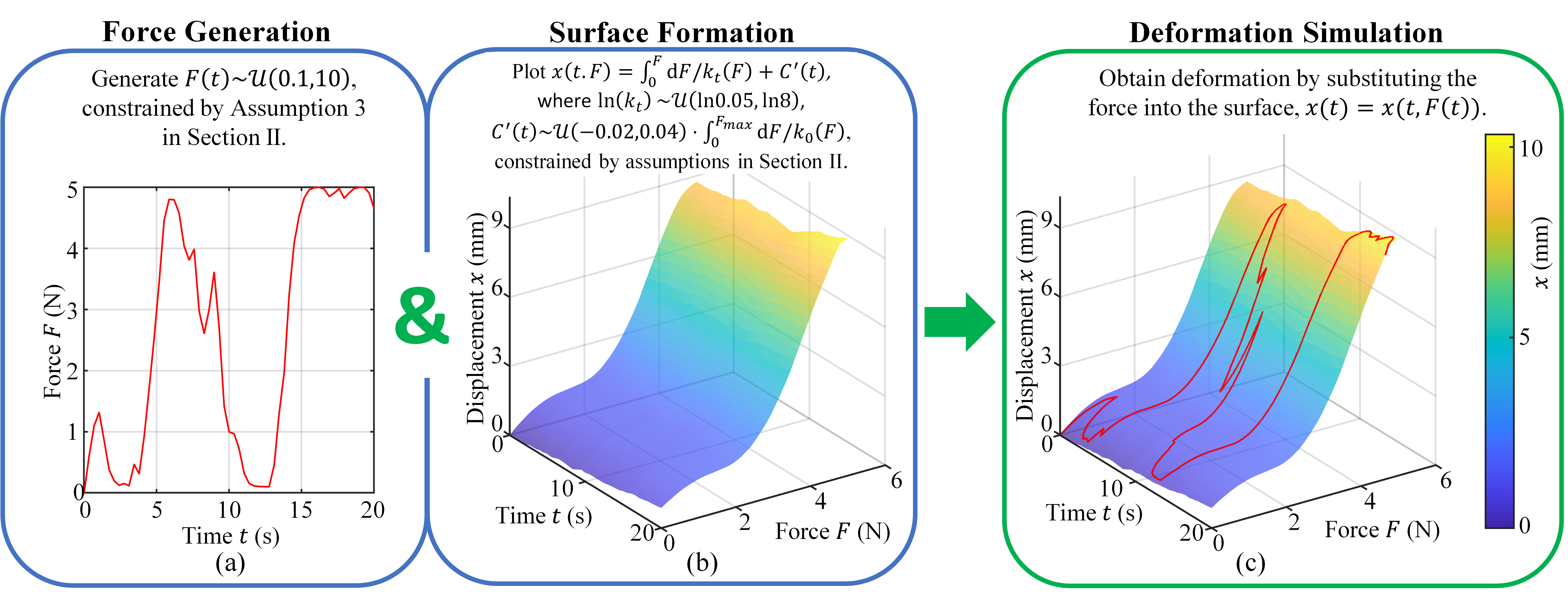}
\caption{{Random data sampling process. (a) Generated grasping force $ F(t) $. (b) Generated surface $ x(t, F) $. (c) Simulated deformation $ x(t) $.}}
\label{fig:sample generating}
\end{figure*}

{In our experiment, the training data were generated by randomly sampling 8,000 grasping processes from this system model based on an empirical distribution, and an additional 2,000 sets (drawn independently but sharing the same distribution) were used for validation. Each dataset simulates a possible grasping process lasting} \( t = \text{20} \, \text{s} \), with the following mechanisms for generating the various physical quantities:

\begin{itemize}
    \item The loading force curve \( F(t) \) during the grasping process does not have a fixed form, so it is randomly generated, ensuring it satisfies Assumption \ref{assumption3} from Chapter \ref{sec:chapter2}. {Empirically, the grasping force appears to be approximately uniformly distributed in the range of} \(\text{0.1} \, \text{N}\) to \(\text{10} \, \text{N}\), so we adopt a uniform distribution as an empirical approximation for \( F \). An example of a randomly generated result is shown in Fig.~\ref{fig:sample generating}(a).
    
    \item The generalized stiffness function \( k(t, \mu(t), F) \) is randomly generated. Since both \( t \) and \( \mu(t) \) affect the stiffness curve in practice, we do not distinguish their individual effects, but ensure the total change is slow and bounded. We generate \( k(t, \mu(t), F) = k_t(F) \) by first generating a function \( k_0(F) \), ensuring \( 0 < k_0(F) < k_m \), and then generating a slowly drifting function \( k_t(F) \) based on \( k_0(F) \), while ensuring that the generated result satisfies Assumption~\ref{assumption6}. 
    
    There are multiple ways to generate the bivariate function \( k_t(F) \), but it is important to note that a simple uniform distribution does not align with empirical observations, as it would result in a low probability of selecting objects with small stiffness. In our scenario, considering Assumption~\ref{assumption2} and real-world objects, the stiffness has an upper bound of \( \text{8} \, \text{N/mm} \) and a lower bound of \( \text{0.05} \, \text{N/mm} \). Therefore, we model the logarithm of stiffness as uniformly distributed, i.e., \( \ln k_t(F) \sim \mathcal{U}(\ln(0.05), \ln(8)) \), which better reflects the real-world distribution of various object stiffness levels.
    
    \item Based on Assumption \ref{assumption4} from Chapter \ref{sec:chapter2}, we randomly generate a slowly changing zero-force drift function \( C(t, \mu(t)) = C'(t) \). In particular, \( C'(t) \) is sampled from a uniform distribution \( \mathcal{U}(-0.02,\; 0.04)\cdot\int_0^{F_{\max}}\frac{dF}{k_0(F)} \), where the scaling factor \( \int_0^{F_{\max}}\frac{dF}{k_0(F)} \) represents the typical deformation magnitude of the object. This design choice is motivated by empirical observations that the magnitude of the zero-force drift is generally much smaller than the object's deformation during the grasping process, and for common viscoelastic materials, inward compression is typically more pronounced than outward expansion, thereby ensuring that the generated \( C'(t) \) reflects realistic behavior.

    \item By substituting \( k_t(F) \) and \( C'(t) \) into \eqref{eq4 nonideal int}, we obtain a surface, as illustrated in Fig.~\ref{fig:sample generating}(b):
    \begin{equation}
    x(t, F) = \int_0^F \frac{\mathrm{d}F}{k_t(F)} + C'(t) \label{eq19 data generating}
    \end{equation}
    
    \item By substituting the previously generated \( F(t) \) into \eqref{eq19 data generating}, we calculate the deformation \( x(t) = x(t, F(t)) \) of the object during the simulated grasping process, as shown in Fig.~\ref{fig:sample generating}(c).

\end{itemize}

Given our hardware conditions, we select a control period \( T = \text{0.01} \, \text{s} \), with 2,000 time steps. The functions generated earlier are sampled at \( \text{0.01} \, \text{s} \) intervals, and we obtain input sequences \( [F(t), x(t)] \) and target output sequences \( k(t, \mu(t), F(t)) = k_t(F(t)) \). By repeating the process described above, multiple sets of simulated grasping data sequences can be obtained. Each randomly generated function has an intuitive physical meaning, and thus can be randomly generated according to the approximate range and distribution characteristics of the various properties of the objects being grasped in real-world scenarios, such as grasp force, zero-force drift, constant-force drift, and stiffness. The distribution of these properties can follow empirical distributions, and random functions can be obtained in various ways, such as selecting characteristic points and performing interpolation.

Then train the network using the following hyperparameters:
\begin{itemize}
    \item \( \text{Epoch} = 300 \)
    \item \( \text{Batch size} = 24 \)
    \item \( \text{Optimizer} = \text{Adam} \)
    \item \( \text{Learning Rate} = 0.0001 \)
\end{itemize}

The data sources for the training and validation process have been introduced earlier in this chapter, while the testing process requires deploying the trained network into real force control tasks, with results presented in Chapter \ref{sec:chapter5}. Here, we show the performance of the LSTM-based stiffness estimator on the validation set and compare it with LSM.

\begin{figure}[t]
\centering
\includegraphics[width=8.8cm]{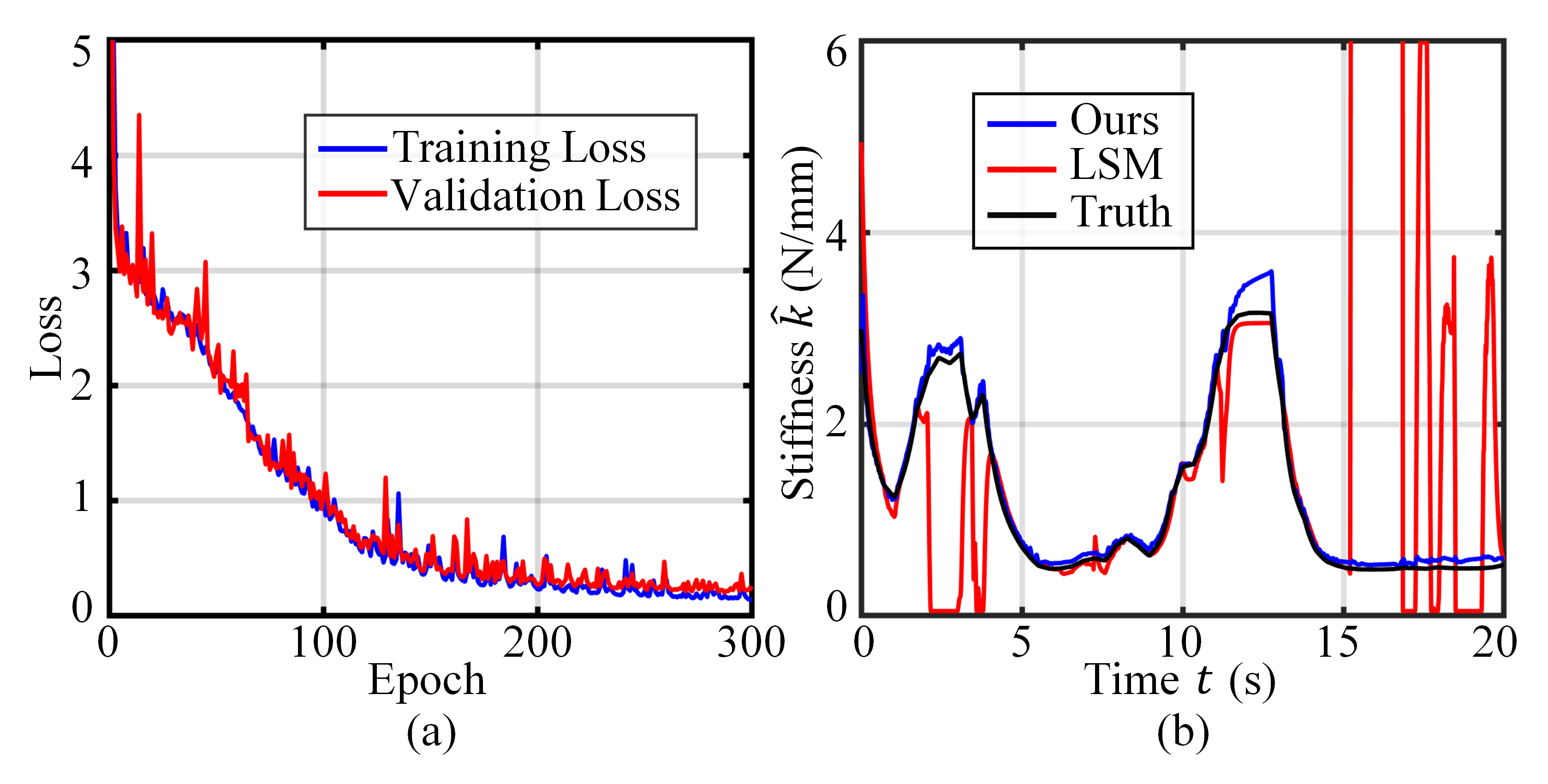}
\caption{{(a) Training and validation loss. (b) Stiffness estimation comparison.}}
\label{fig:validation}
\end{figure}

As shown in Fig.~\ref{fig:validation}(a), the training and validation errors plateau towards the end of training, indicating convergence and no overfitting. The training and validation errors are close, demonstrating good generalization ability of the neural network to unfamiliar grasping data. On the validation set, our method achieved a mean error of \( \text{Mean}( \text{loss} ) = \text{0.144} \) and variance \( \text{Var}( \text{loss} ) = \text{0.0537} \), while LSM had a mean error of \( \text{Mean}( \text{loss} ) = \text{142.4} \) and variance \( \text{Var}( \text{loss} ) = \text{54265.1} \). This shows that our method provides smaller average errors and higher estimation accuracy compared to the traditional LSM, and the lower variance indicates better robustness.

Using the data shown in Fig.~\ref{fig:sample generating}, the stiffness estimation results are shown in Fig.~\ref{fig:validation}(b). Our method provides a good fit and does not exhibit large errors at specific time points, while LSM tends to make significant errors at turning points or inflection points in the force curve, demonstrating better robustness. Moreover, after the initial few time steps, the stiffness estimate converges to the true value and remains accurate, eliminating the need for prolonged probing times.

This section demonstrates a high-precision, short-probing-time online generalized stiffness estimator that can be applied in online optimization of controller parameters, laying the foundation for achieving high-precision and short-probing-time force tracking control.

\section{Design of the Force Tracking Controller}
\label{sec:chapter4}
\begin{figure}[t]
    \centering
    \includegraphics[width=8.8cm]{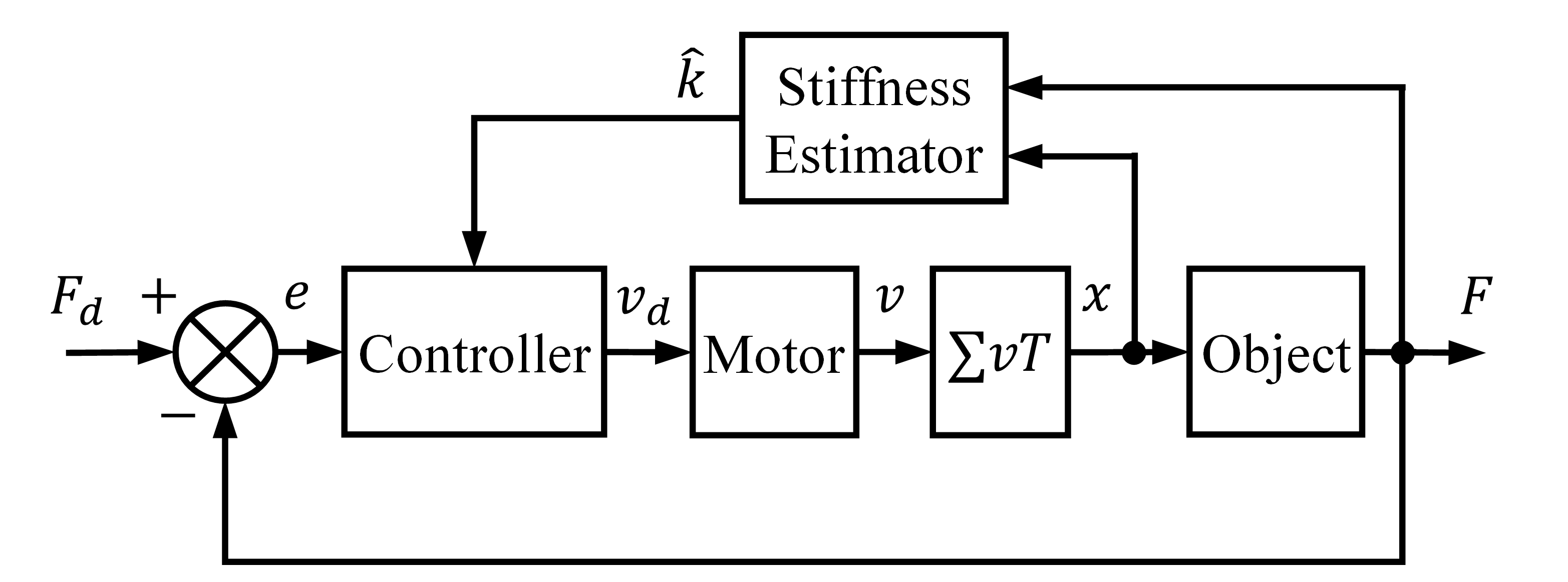}
    \caption{Control block diagram.}
    \label{fig:block_diagram}
\end{figure}

This section designs an adaptive force tracking controller with online parameter tuning based on generalized stiffness, and analyzes the relationship between its stability, convergence speed, accuracy, and the generalized stiffness estimation. Considering that most robotic manipulators are equipped with high-frequency and precise speed controllers, we design the force tracking controller based on this. Without loss of generality, we implement adaptive parameter tuning based on generalized stiffness on a common PI controller. The system control block diagram is shown in Fig.~\ref{fig:block_diagram}.

Assume that the desired force curve to be tracked is \( F_d(t) \), the actual velocity at time \( t \) is \( v(t) \), and the desired velocity is \( v_d(t) \). The control law is as follows:
\begin{equation}
    v_d(t+T) = \frac{K_P}{\hat{k}(t)} e(t) + \frac{K_I}{\hat{k}(t)} \sum_{q=0}^{\frac{t}{T}} e(qT)
    \label{eq:control_law}
\end{equation}
where \( e(t) = F_d(t) - F(t) \) is the force tracking error, \( \hat{k}(t) > 0 \) is the output of the generalized stiffness estimator, and ideally, we want \( \hat{k}(t) = k(t, \mu(t), F(t)) \), where \( k(t, \mu(t), F(t)) \) is the true value of the generalized stiffness. \( K_P \) and \( K_I \) are constant parameters.

The generalized stiffness reflects the ratio of small displacement to force variation. A larger stiffness means that the displacement needed to control the motor when there is force tracking error is smaller. Therefore, the PI coefficients in the traditional PI controller are divided by the generalized stiffness to achieve automatic parameter tuning. Below, we will analyze the effect of \( \hat{k}(t) \) on the system's stability and convergence speed.

Although the velocity tracking performance of the robotic manipulator is not ideal, assume it is stable, its error being a bounded small quantity, i.e.,
\begin{equation}
    v(t) = v_d(t) + \epsilon_v(t), \quad |\epsilon_v(t)| < \delta_v
    \label{eq:velocity_error}
\end{equation}

When the control period \( T \) is short, the relationship between velocity and displacement is approximately:
\begin{equation}
    x(t+T) - x(t) = v(t+T)T = (v_d(t+T) + \epsilon_v(t+T))T
    \label{eq:displacement_velocity_relation}
\end{equation}

Substituting the assumptions from \eqref{eq9 Positive Continuous Bounded desired Force}, \eqref{eq10 Zero Force Drift is Slow and Bounded}, and~\eqref{eq13 Constant Force Drift is Slow} into \eqref{eq7 High Control Frequency}, we get:
\begin{align}
    &x(t+T) - x(t) \notag\\
    =&\frac{1}{k(t, \mu(t), F(t))} \left( F(t+T) - F(t) \right) + \epsilon_{kF}(t) + \epsilon_C(t) \notag\\
    =&\frac{1}{k(t, \mu(t), F(t))} \left( \epsilon_{F_d}(t) - (e(t+T) - e(t)) \right) \notag\\
    +& \epsilon_{kF}(t) + \epsilon_C(t)
    \label{eq:left_side_displacement}    
\end{align}

Substituting \eqref{eq:left_side_displacement} and~\eqref{eq:control_law} into the left and right sides of \eqref{eq:displacement_velocity_relation}, respectively, and then multiplying both sides by \( k(t, \mu(t), F(t)) \), we obtain:
\begin{align}
    &P(t)e(t) + I(t)\sum_{q=0}^{\frac{t}{T}} e(qT) + (e(t+T) - e(t)) \notag\\
    =& \epsilon_{F_d}(t) + k(t, \mu(t), F(t)) (\epsilon_{kF}(t) + \epsilon_C(t) - T\epsilon_v(t+T))\notag\\
    =& d(t)
    \label{eq:system_equation}
\end{align}
where \( P(t) = K_P T k(t, \mu(t), F(t))/\hat{k}(t)  \) and \( I(t) = K_I T k(t, \mu(t), F(t))/\hat{k}(t) \). \( d(t) \) is also a bounded small quantity because:
\begin{align}
    &|d(t)| \leq |\epsilon_{F_d}(t)| \notag\\
    +& k(t, \mu(t), F(t)) (|\epsilon_{kF}(t)| + |\epsilon_C(t)| + T|\epsilon_v(t+T)|) \notag\\
    <& \delta_{F_d} + k_m (\delta_{kF} + \delta_C + T\delta_v) = \delta_d
    \label{eq:bounded_error}    
\end{align}

Let the accumulated error be \( s(t) = \sum_{q=0}^{t/T} e(qT) \), then we have:
\begin{equation}
    s(t+T) = s(t) + e(t+T)
    \label{eq:accumulated_error}
\end{equation}

Equation~\eqref{eq:system_equation} can be rewritten as:
\begin{equation}
    P(t) e(t) + I(t) s(t) + (e(t+T) - e(t)) = d(t)
    \label{eq:revised_equation}
\end{equation}

Let the system state vector be \( a(t) = \begin{bmatrix}s(t) \\ e(t)\end{bmatrix} \), then \eqref{eq:accumulated_error} and~\eqref{eq:revised_equation} can be written in matrix form as:
\begin{equation}
    a(t+T) = \begin{bmatrix}1 - I(t) & 1 - P(t) \\ -I(t) & 1 - P(t)\end{bmatrix} a(t) + \begin{bmatrix}d(t) \\ d(t)\end{bmatrix}
    \label{eq:matrix_form}
\end{equation}

Let \( A(t) = \begin{bmatrix}1 - I(t) & 1 - P(t) \\ -I(t) & 1 - P(t)\end{bmatrix} \) and \( D(t) = \begin{bmatrix}d(t) \\ d(t)\end{bmatrix} \), then \eqref{eq:matrix_form} becomes:
\begin{equation}
    a(t+T) = A(t) a(t) + D(t)
    \label{eq:linear_system}
\end{equation}
Where the input of the system \( \| D(t) \| = \sqrt{2} \, | d(t) | < \sqrt{2} \, \delta_d \) is a bounded small quantity.

\begin{figure}[t]
    \centering
    \includegraphics[width=8.8cm]{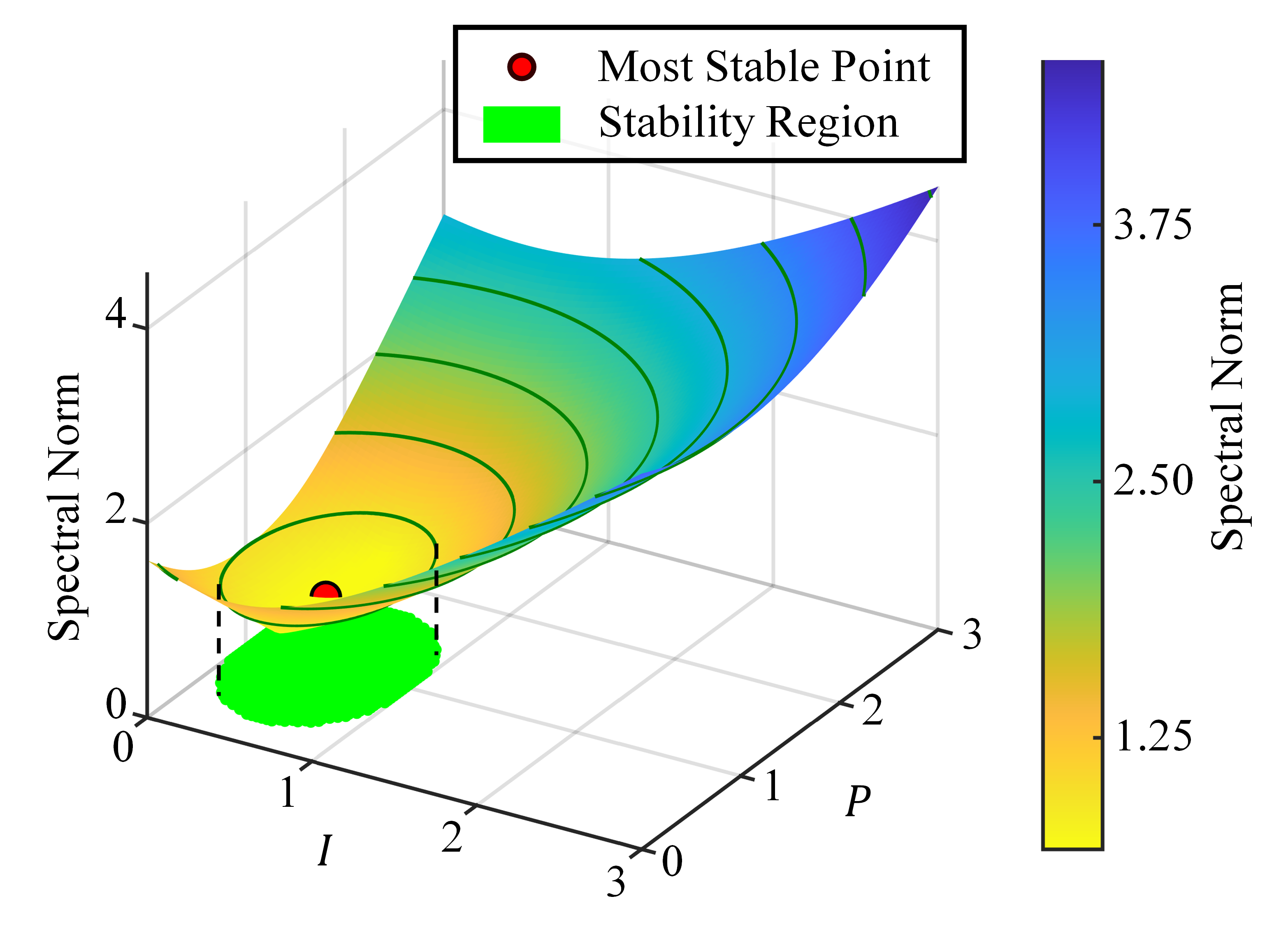}
    \caption{Spectral norm of the system matrix.}
    \label{fig:spectral_norm}
\end{figure}

The spectral norm of the system matrix \( \| A \| \) as a function of \( I \) and \( P \) is shown in Fig.~\ref{fig:spectral_norm}. When \( I = \text{0.5} \) and \( P = \text{1.0} \), \( |A| \) reaches its minimum value of 0.707, and in a region near this point, \( \| A \| < \text{1} \). If we can ensure that \( I(t) \) and \( P(t) \) always remain within this stable region, the matrix \( A \) will decay rapidly, and the error \( e(t) \) will approach zero quickly, ensuring system stability. Moreover, when \( I = \text{0.5} \) and \( P = \text{1.0} \), the convergence speed is the fastest.

Assuming that the generalized stiffness estimation has no bias, we have \( P(t) = K_P T \) and \( I(t) = K_I T \). To place the system at the most stable point where \( I = \text{0.5} \) and \( P = \text{1.0} \), we need to set \( K_P = 1/T \) and \( K_I = 1/2T \). After determining these two constant parameters, let the ratio of the generalized stiffness estimate to the true value be \( \eta(t) = \hat{k}(t)/k(t, \mu(t), F(t)) \), then the system matrix becomes:
\begin{equation}
    A(t) = \begin{bmatrix}1 - \frac{1}{2\eta(t)} & 1 - \frac{1}{\eta(t)} \\ -\frac{1}{2\eta(t)} & 1 - \frac{1}{\eta(t)}\end{bmatrix}
    \label{eq:system_matrix}
\end{equation}

The condition for system convergence, i.e., \( \| A(t) \| < 1 \), is:
\begin{equation}
    0.634 < \eta(t) < 2.366
    \label{eq:convergence_condition}
\end{equation}

This section proposes a force tracking strategy that considers generalized stiffness. The analysis shows that in order to guarantee system convergence, the gains of the PI controller must indeed be adjusted according to the generalized stiffness estimate, and there is a certain tolerance for the estimation. As long as \( \eta(t) \) lies within the interval \( (\text{0.634}, \text{2.366}) \), convergence will be guaranteed. Moreover, the closer \( \eta(t) \) is to \( \text{1} \), the faster the convergence speed. Therefore, the accuracy of the generalized stiffness estimate is a key factor affecting force tracking performance. In experiments, the performance of the generalized stiffness estimator can be reflected by the force tracking performance.

\section{Experimental Verification}
\label{sec:chapter5}
This section introduces the experimental platform used for grasping, where the adaptive force tracking method proposed in the previous sections is implemented. The performance of the proposed method is compared with the method from \cite{ref17}, validating the excellent performance achieved by the PI controller with the generalized stiffness estimator for online parameter tuning. This includes the adaptability in terms of accuracy and probing time. Finally, the application of the proposed force tracking algorithm in a complete grasping task is demonstrated.

\subsection{Experimental Setup}
\begin{figure}[t]
    \centering
    \includegraphics[width=8.8cm]{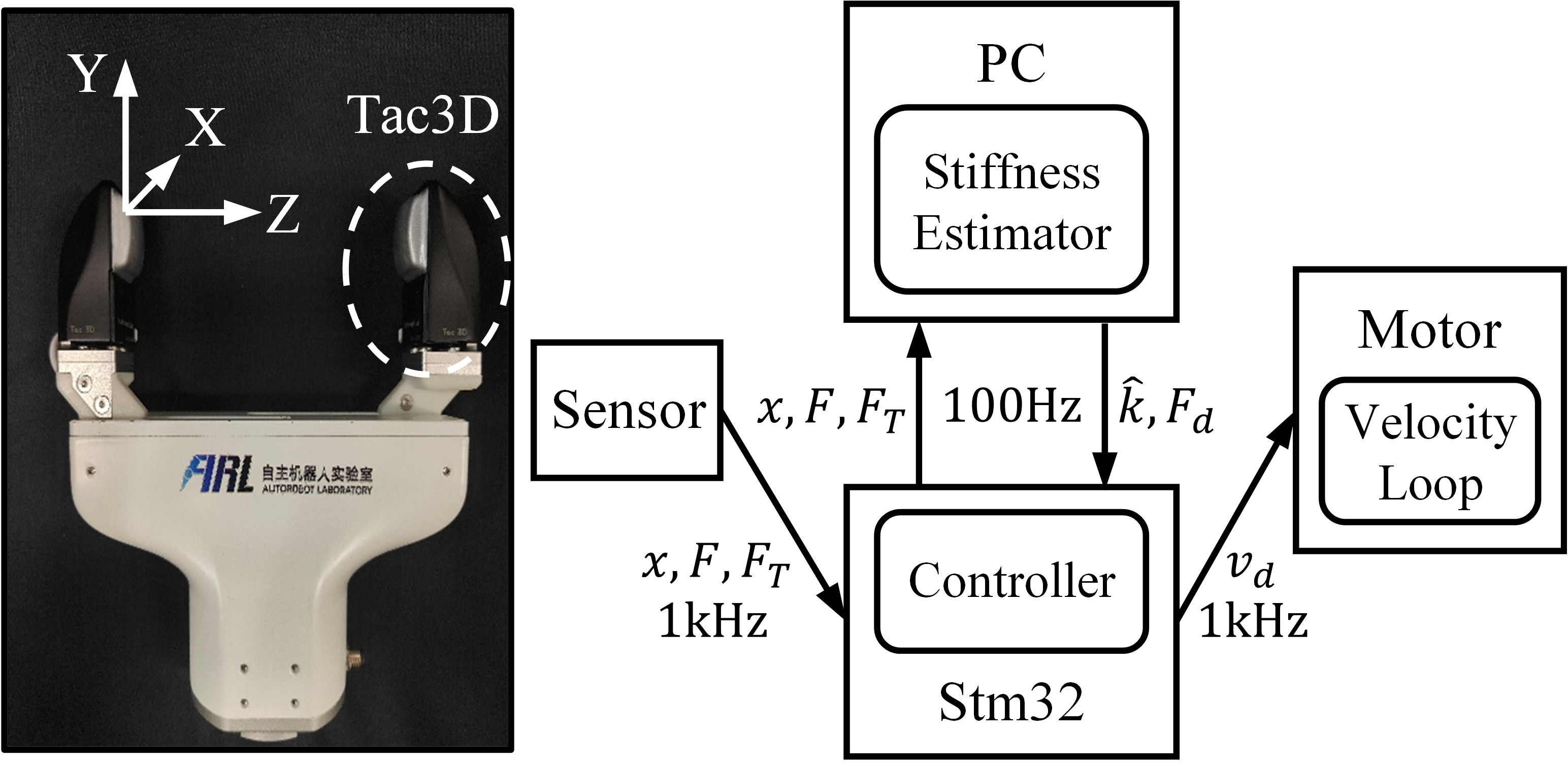}
    \caption{Robotic gripper and communication diagram.}
    \label{fig:comm_diagram}
\end{figure}

The force tracking task is completed by a single-degree-of-freedom two-finger robotic gripper, as shown in Fig.~\ref{fig:comm_diagram}. The two fingers can move in reverse synchrony along the \(Z\)-axis. The gripper is equipped with Tac3D tactile sensors and encoders, capable of measuring the force in the \(Z\)-direction, \( F \), the force in the \(X-Y\) plane, \( F_T \), and the position, \( x \). In fact, although this article focuses on the grasping force tracking, numerous studies have demonstrated that vision-based tactile sensors are capable of measuring more complex force information and guiding the setting of target grasping forces \cite{ref20,ref21}. By combining these capabilities, a comprehensive grasping force control system can be achieved. The gripper has an STM32 microcontroller, which receives force sensor data, \( F \), and encoder readings, \( x \), at a frequency of \( \text{1} \, \text{kHz} \) and sends speed control commands, \( v_d \), to the motors. {The host computer, a Personal Computer (PC) powered by an AMD Ryzen 7 6800H CPU without GPU acceleration, reads the force and position information from the microcontroller at a frequency of  $ \text{100} \, \text{Hz} $, performs real-time stiffness estimation and control computations, and transmits the estimated generalized stiffness $ \hat{k} $ and target force $ F_d $ in real time}. The communication diagram is shown in Fig.~\ref{fig:comm_diagram}.

The experimental procedure is performed in four steps:

\begin{enumerate}
    \item Initially, the object to be grasped is placed on a flat table, positioned between the two fingers of the gripper.
    \item The gripper slowly closes, and the force reading reaches \( \text{0.5} \, \text{N} \), at which point the force tracking process begins.
    \item The target force curve is given, and the force tracking controller is used to regulate the actual grasping force.
    \item Different objects are selected for the experiment, and the above process is repeated, with the experimental data recorded by the host computer.
\end{enumerate}

\begin{figure*}[htbp]
    \centering
    \includegraphics[width=16cm]{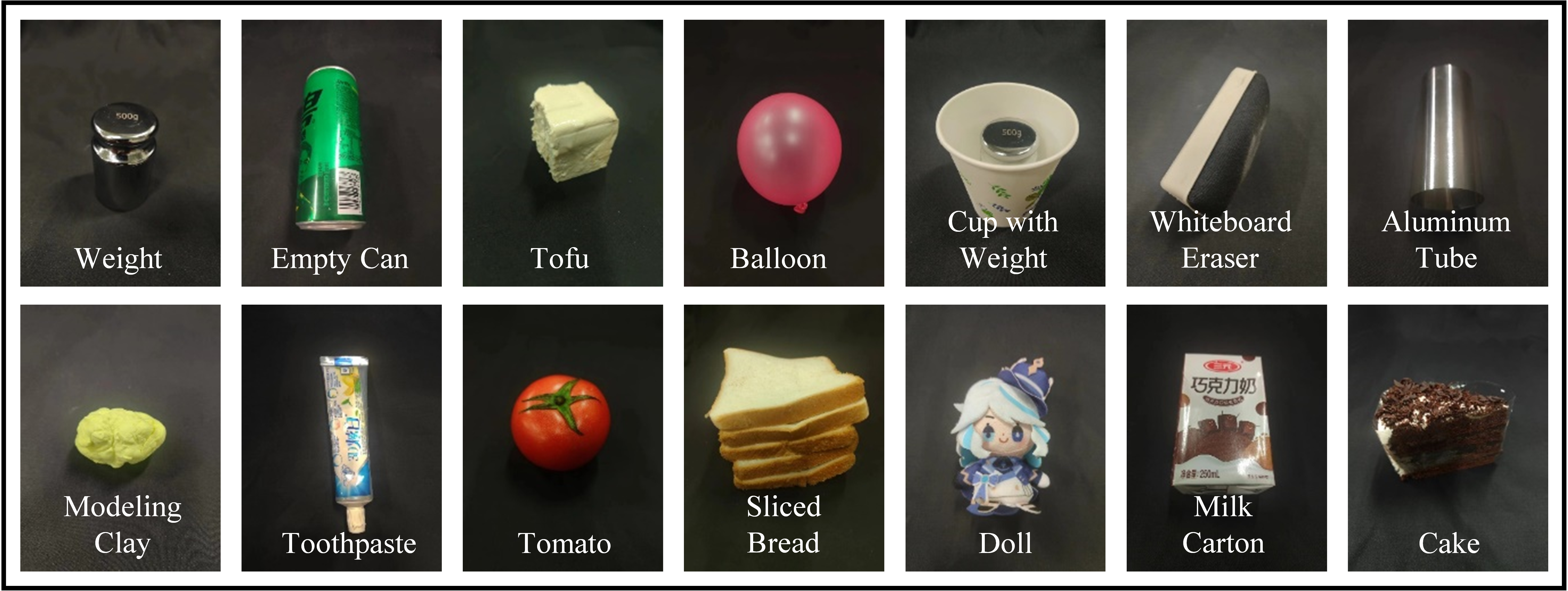}
    \caption{Objects to be grasped in the experiment.}
    \label{fig:grasp_objects}
\end{figure*}

To compare with existing research performance, experiments are conducted using both the proposed method and the MiFREN method from \cite{ref17}. Excluding methods with significant adaptation deficiencies for non-ideal objects, this method is currently the state-of-the-art adaptive force tracking algorithm. In the experiment, the target grasping force curve \( F_d(t) \) is given. Since both methods require parameter adjustment based on the initial probing of the object, the aim of this experiment is to verify the probing time required for both methods to achieve optimal performance and the asymptotic accuracy achieved after sufficient probing. Therefore, the target force curve is set as a periodic trapezoidal load. The objects used in the experiment are shown in Fig.~\ref{fig:grasp_objects}, covering most common types. These include weights, empty cans, tofu, and balloons, representing objects with a range of stiffness from hard to soft. Paper cups with weights, whiteboard erasers, and aluminum tubes are typical variable-stiffness objects whose stiffness changes significantly with different applied forces. Additionally, some objects exhibit mechanical behaviors that deviate from ideal elasticity and cannot be accurately modeled using traditional stiffness concepts. For these objects, generalized stiffness must be introduced to capture their characteristics. Examples include modeling clay and toothpaste, which are typical plastic materials; tomatoes and bread, which exhibit pronounced viscoelasticity; and dolls, cakes, and milk cartons, which have complex and poorly understood contact mechanics.

\subsection{Experimental Data Analysis}
We first analyze the experimental results of two force tracking strategies using a paper cup with weight as an example.

\begin{figure}[t]
    \centering
    \includegraphics[width=8.8cm]{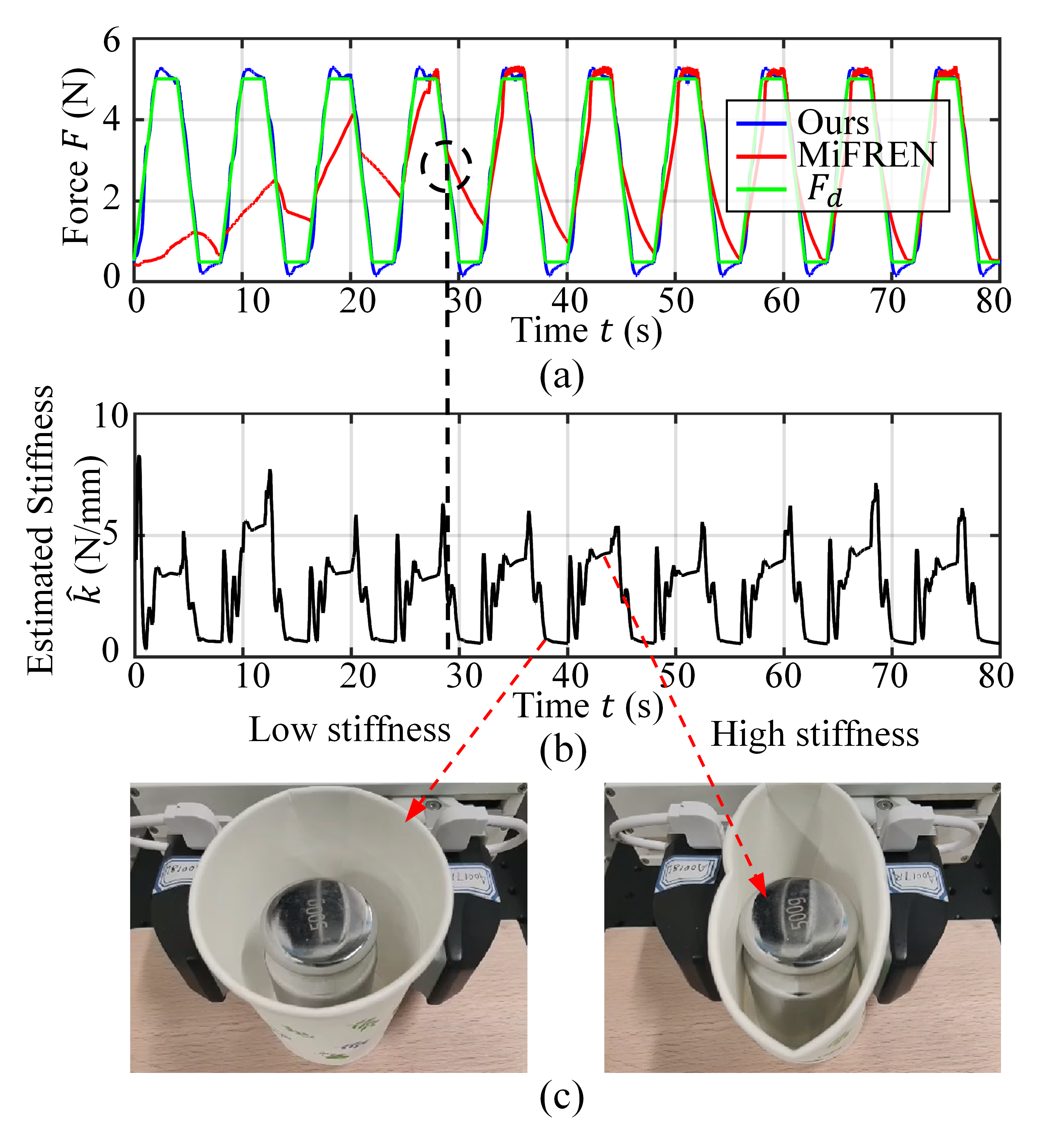}
    \caption{{Experimental results for the paper cup with weight. (a) Curves of the actual force and the target force $F_d$. (b) Stiffness estimation results. (c) Object deformation after force application.}}
    \label{fig:paper_cup_experiment}
\end{figure}

As shown in Fig.~\ref{fig:paper_cup_experiment}(c), when the paper cup with weight is subjected to large grasping forces, the stiffness increases sharply due to the cup wall tightly pressing against the weight. Therefore, the force tracking strategy must have the ability to quickly adapt to such changes. Observing the force tracking curve in Fig.~\ref{fig:paper_cup_experiment}(a) at around \( t \approx \text{29} \, \text{s} \), when the grasping force decreases, the object's stiffness suddenly decreases, and the MiFREN algorithm shows a sudden increase in error, indicating its poor adaptability to such stiffness changes, while our method maintains a small error, demonstrating better adaptability. This is because, as shown in Fig.~\ref{fig:paper_cup_experiment}(b), the stiffness estimator immediately adjusts the stiffness estimate when the object's stiffness undergoes a sudden change, allowing the force tracking controller parameters to be rapidly adjusted to appropriate values, ensuring good tracking performance.

\begin{figure}[t]
    \centering
    \includegraphics[width=8.8cm]{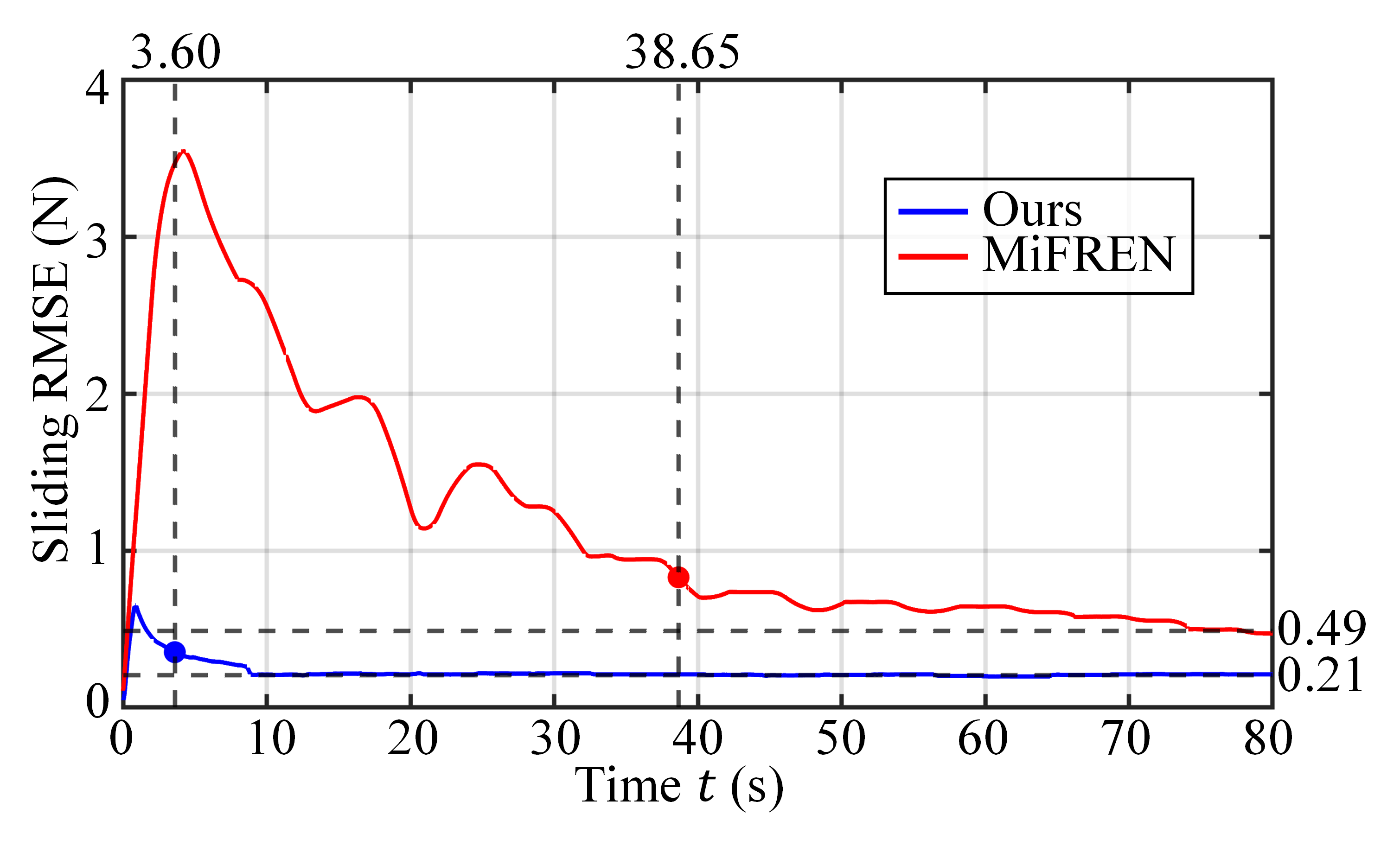}
    \caption{{Comparison of sliding RMSE for the cup with weight.}}
    \label{fig:rmse_comparison}
\end{figure}

Next, we quantitatively analyze the accuracy and probing time of the two force tracking algorithms. In Fig.~\ref{fig:paper_cup_experiment}(a), for any given time \( t \), we calculate the Root Mean Square Error (RMSE) of the force tracking performance over the previous cycle. If \( t \) is less than one cycle, we compute the RMSE from \( \text{0} \) to \( t \). This is referred to as the sliding RMSE. The sliding RMSE of both methods is shown in Fig.~\ref{fig:rmse_comparison}. As seen, when \( t \) is large enough, the sliding RMSE tends to a constant value, and the horizontal asymptote represents the asymptotic error, which reflects the final accuracy achievable after the algorithm has run for a sufficient amount of time on the same object. The time at which the sliding RMSE converges within \( \text{70}\% \) of its asymptotic error is referred to as the probing time, which reflects the time required for the algorithm to reach a high accuracy. For the paper cup with weight, the asymptotic error of our method is \( \text{0.21} \, \text{N} \), while MiFREN's asymptotic error is \( \text{0.49} \, \text{N} \). The probing time for our method is \( \text{3.60} \, \text{s} \), whereas MiFREN's probing time is \( \text{38.65} \, \text{s} \). This indicates that the method proposed in this paper achieves higher force tracking accuracy and significantly shorter probing time than MiFREN for this object.

\begin{figure}[t]
    \centering
    \includegraphics[width=8.8cm]{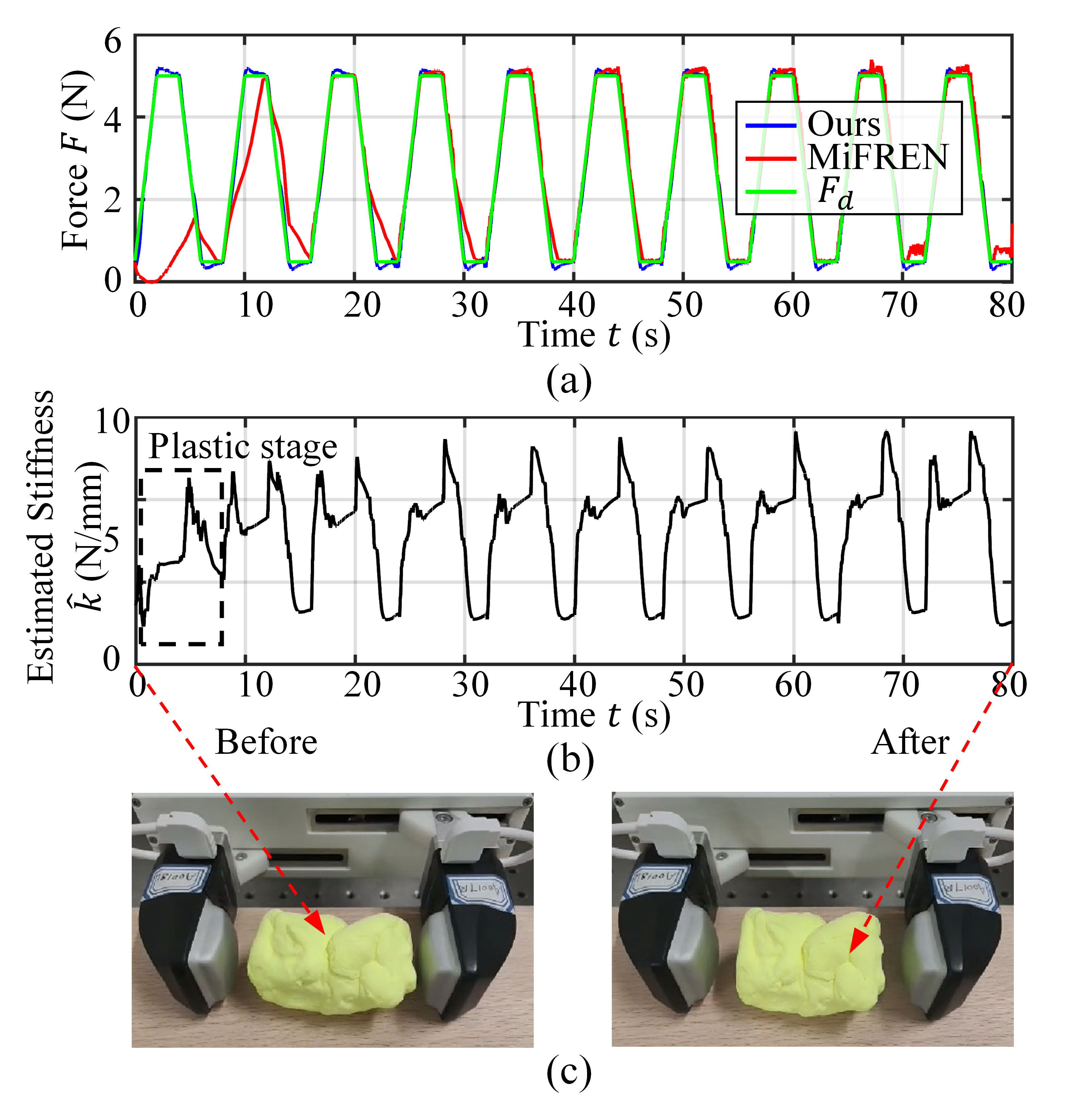}
    \caption{{Experimental results for the modeling clay. (a) Curves of the actual force and the target force $F_d$. (b) Stiffness estimation results. (c) Object deformation after force application.}}
    \label{fig:clay_experiment}
\end{figure}

In addition to variable-stiffness objects, we also demonstrate the performance of both force tracking algorithms on plastic objects, using modeling clay as an example. As shown in Fig.~\ref{fig:clay_experiment}(c), clay shortens in length after grasping, exhibiting plasticity. The traditional stiffness concept is insufficient to describe the behaviors of such objects. However, the generalized stiffness estimation results shown in Fig.~\ref{fig:clay_experiment}(b) effectively reflect this behavior. In the first grasping force loading cycle, the generalized stiffness estimation exhibits a different pattern compared to the subsequent cycles. This is because the clay initially undergoes a plastic compression stage, and after compression, it no longer rebounds, so its physical properties before and after compression differ, resulting in a change in the behavior of generalized stiffness during periodic force applications.

\begin{figure}[t]
    \centering
    \includegraphics[width=8.8cm]{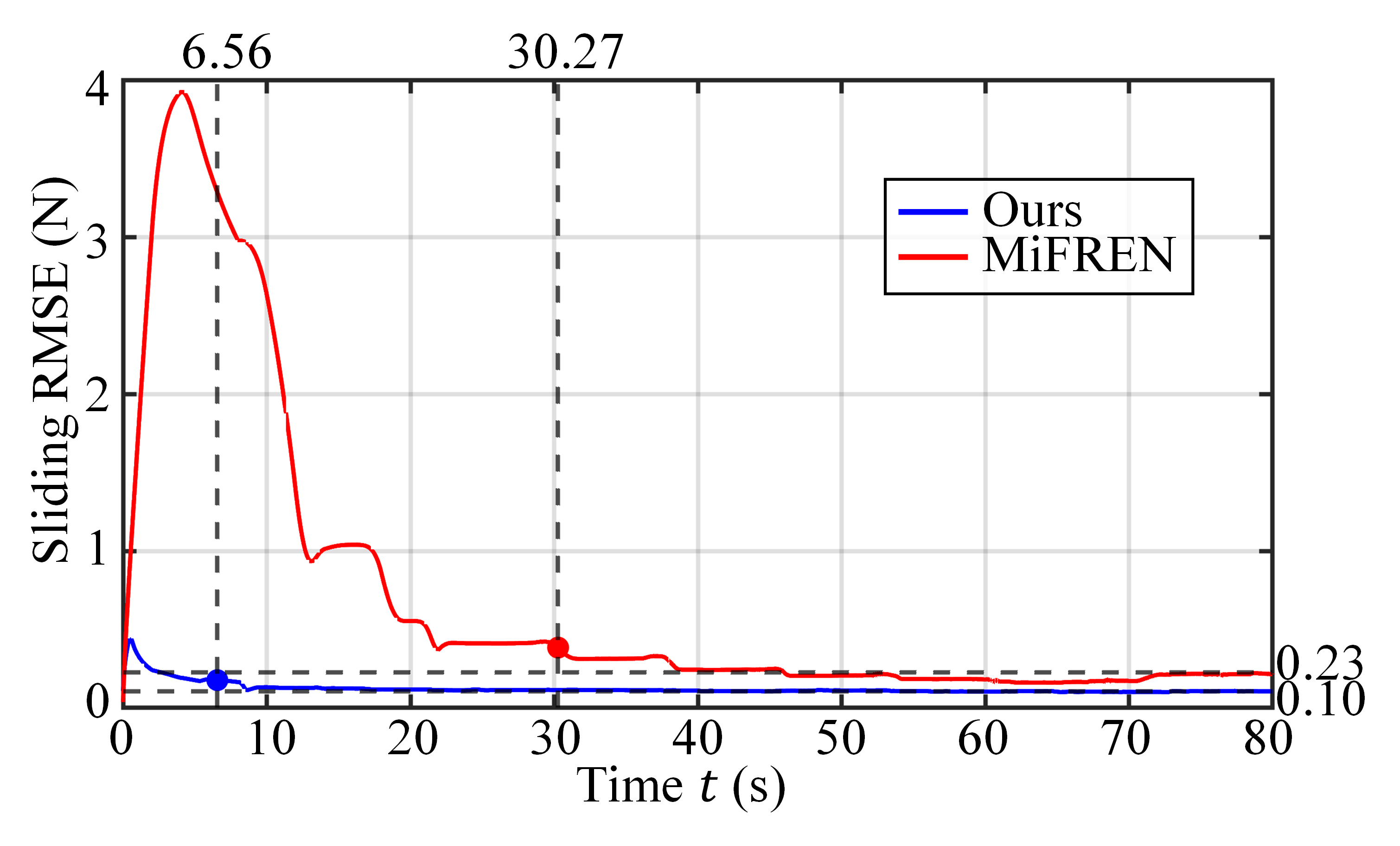}
    \caption{{Comparison of sliding RMSE for the modeling clay.}}
    \label{fig:rmse_comparison_clay}
\end{figure}

As shown in Fig.~\ref{fig:rmse_comparison_clay}, the comparison of the sliding RMSE for the clay further demonstrates that our method outperforms MiFREN in terms of both asymptotic error and probing time. Additionally, in the early stage (\(\text{0-8} \, \text{s}\)), the sliding RMSE of MiFREN reaches nearly \(\text{4} \, \text{N}\), indicating that it does not adapt well to the plastic stage of clay, while our method maintains a sliding RMSE of less than \(\text{0.5} \, \text{N}\), ensuring good force tracking accuracy even during the plastic stage.

Finally, we summarize the experimental results for all objects, calculating the asymptotic errors and probing times for each, as shown in Table~\ref{tab:comparison}. On average, our method achieves an asymptotic error of \(\text{0.16} \, \text{N}\), about one-third of MiFREN's \(\text{0.49} \, \text{N}\). The probing time for our method is only \(\text{3.34} \, \text{s}\), less than one-eleventh of MiFREN's \(\text{38.65} \, \text{s}\). This shows that the proposed method achieves higher force tracking accuracy and shorter probing time across various objects, including elastic, viscoelastic, plastic, and variable-stiffness materials, demonstrating excellent adaptability. For most objects, our method achieves a smaller asymptotic error, with MiFREN showing a slight advantage only on specific objects such as empty cans, tomatoes, and milk cartons. Furthermore, MiFREN exhibits significantly higher errors for aluminum tubes and balloons, which may be due to the fact that these are variable-stiffness objects, and their stiffness is significantly lower than that of other objects under smaller grasping forces, which MiFREN fails to adapt to in a limited time. In contrast, our method achieves similar asymptotic errors across all objects, indicating that the generalized stiffness estimator is accurate and effectively adjusts the controller parameters to suitable values. Considering the probing time, our method requires significantly less time than MiFREN for all objects, showing that it can explore and adapt to the object's characteristics much more quickly, making it more suitable for rapid grasping.

{To further understand the performance of our proposed method, we analyze the primary sources of error in our system:}

\begin{itemize}
    \item \textbf{Limited performance of mechanical actuators}: Our force control is based on the built-in velocity loop of the robotic manipulator's motors, which introduces velocity control errors. As a result, the manipulator cannot perfectly follow our motion commands. More precise actuators would help reduce the existing errors.
    
    \item \textbf{Theoretical limitations of the control model}: Since our force control relies on a PI controller, as shown in Figure~\ref{fig:spectral_norm}, the system matrix has a minimum spectral norm of 0.707. This implies that the system still requires several time steps to converge the error close to zero. Adopting a control strategy with faster convergence than PI control as the underlying framework would help accelerate the error convergence.
    
    \item \textbf{Errors in stiffness estimation}: Stiffness estimation is performed at a frequency of 100 Hz and is based on predicting future stiffness changes using historical information. Abrupt changes in the object's stiffness (e.g., a paper cup with weight) can cause the estimated stiffness to deviate from its true value, leading to a deviation of the system matrix's spectral norm from its minimum and slowing down the convergence speed. Enhancing stiffness estimation algorithms may help speed up error convergence.
\end{itemize}

\begin{table}[htbp]
\caption{Comparison Between Two Methods}
\label{tab:comparison}
\centering
\resizebox{\columnwidth}{!}{%
\begin{tabular}{|c|c|c|c|c|}
\hline
 & \multicolumn{2}{c|}{\textbf{Ours}} & \multicolumn{2}{c|}{\textbf{MiFREN}} \\ \hline
\textbf{Object} & \textbf{\makecell{Asymptotic \\ Error (N)}} & \textbf{\makecell{Probing \\ Time (s)}} & \textbf{\makecell{Asymptotic \\ Error (N)}} & \textbf{\makecell{Probing \\ Time (s)}} \\ \hline
Weight & 0.09 & 1.14 & 0.09 & 30.22 \\ \hline
Empty Can & 0.22 & 0.07 & 0.19 & 78.00 \\ \hline
Tofu & 0.17 & 3.26 & 0.18 & 54.02 \\ \hline
Balloon & 0.24 & 3.32 & 1.55 & 10.63 \\ \hline
Cup with Weight & 0.20 & 3.60 & 0.49 & 38.65 \\ \hline
Whiteboard Eraser & 0.17 & 3.12 & 0.20 & 46.12 \\ \hline
Aluminum Tube & 0.28 & 3.98 & 1.47 & 11.69 \\ \hline
Modeling Clay & 0.10 & 6.56 & 0.23 & 30.27 \\ \hline
Toothpaste & 0.10 & 8.70 & 0.14 & 37.58 \\ \hline
Tomato & 0.15 & 1.70 & 0.13 & 30.63 \\ \hline
Sliced Bread & 0.18 & 3.97 & 0.45 & 39.49 \\ \hline
Doll & 0.19 & 3.84 & 0.50 & 31.43 \\ \hline
Milk Carton & 0.08 & 1.79 & 0.07 & 45.35 \\ \hline
Cake & 0.11 & 1.66 & 0.16 & 30.01 \\ \hline
\textbf{Average} & \textbf{0.16} & \textbf{3.34} & \textbf{0.42} & \textbf{36.72} \\ \hline
\end{tabular}
}
\end{table}

\subsection{Application Demonstration}
\begin{figure}[t]
    \centering
    \includegraphics[width=8.8cm]{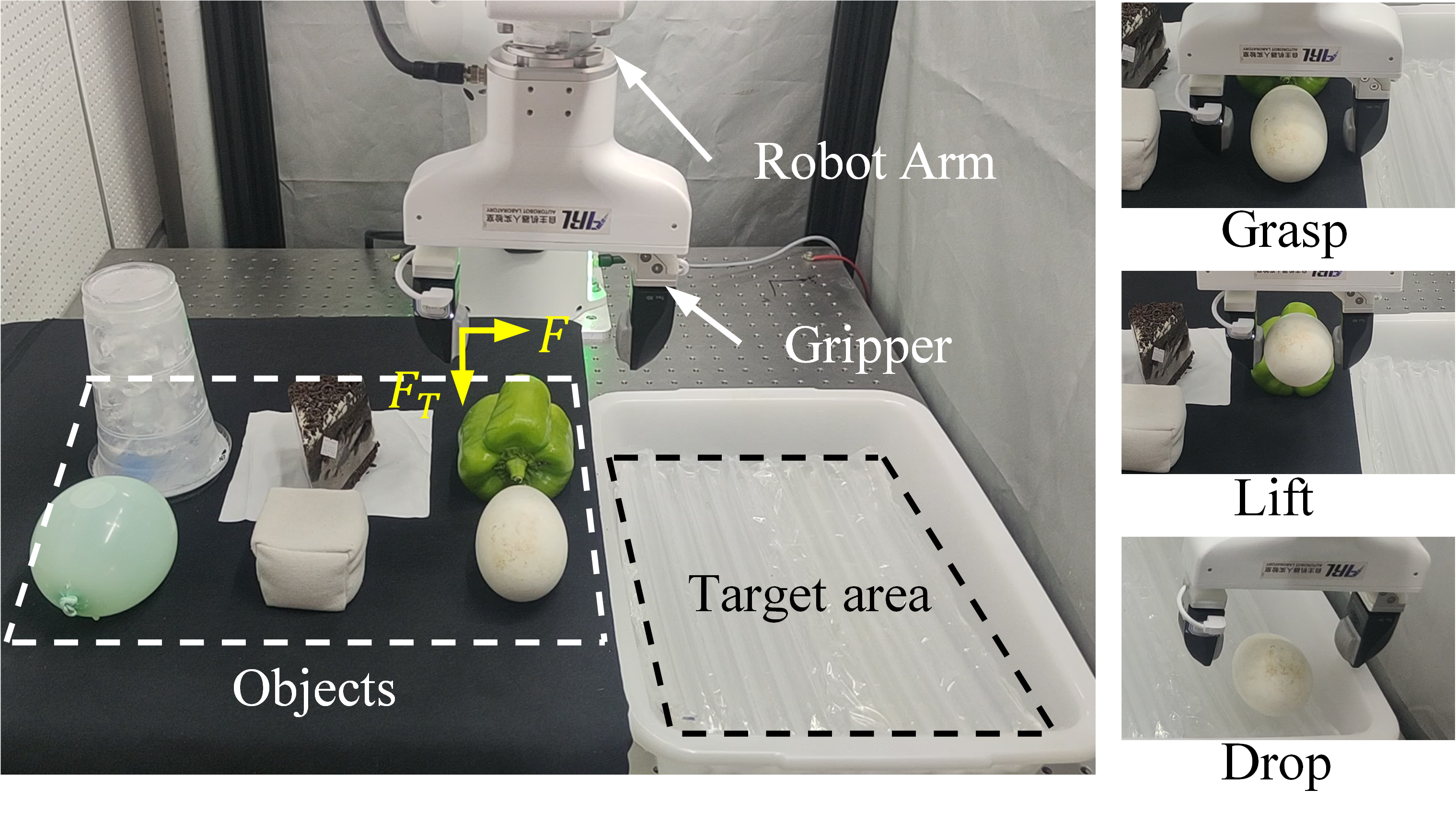}
    \caption{Grasping experimental platform.}
    \label{fig:grasping_platform}
\end{figure}

To demonstrate the effectiveness of the force tracking method proposed in this paper for grasping tasks, the following experimental design is presented. The experimental platform is shown in Fig.~\ref{fig:grasping_platform}. The robotic arm is responsible for lifting the gripper, and the gripper and communication framework are the same as in Fig.~\ref{fig:comm_diagram}. The robotic arm and gripper must complete the task of transporting and collecting several objects, moving them from the platform to a target area, ensuring that the objects are neither damaged nor dropped during the process.The steps are as follows:

\begin{itemize}
    \item The robotic arm moves the gripper so that the object is placed between the gripper fingers.
    \item The gripper slowly closes. When the normal force reaches \( \text{1} \, \text{N} \), it is considered that the object has been contacted. The robotic arm then lifts the object, and the gripper performs force planning and force tracking to ensure that the object does not slip during the subsequent process.
    \item The robotic arm moves the gripper and object near the target area, and then the object is released.
\end{itemize}

In the experiment, objects such as goose eggs, sandbags, water balloons, green peppers, cakes, and ice cups are selected for grasping. Among them, the goose egg has relatively high stiffness but is prone to breakage when subjected to vibration, impact, or large grasping forces, requiring precise control of the actual grasping force. The sandbag contains sand particles that have some fluidity, but when subjected to large pressure, it exhibits rigidity. Moreover, when the sandbag is lifted off the table, the sand loses its support and flows toward the low points, causing the macroscopic shape of the sandbag to change, which presents challenges for force tracking. The stiffness of the water balloon is much smaller than that of the goose egg, and the gripper needs to automatically adapt to this. Green peppers and cakes are fragile objects, and the cake has complex and unknown contact mechanical behaviors, requiring the force tracking algorithm to be highly accurate and adaptive. The ice cup is a variable-stiffness object, with significant differences in stiffness between the cup and the ice cubes, and the internal movement of the ice cubes during the grasping process may further cause changes in the object characteristics. Therefore, the force tracking algorithm must be capable of adapting to these changes in real-time.

As mentioned in Chapter \ref{sec:chapter1}, the control of the grasping force by the gripper consists of force planning and force tracking components. There are various existing methods for force planning \cite{ref6, ref7, ref8, ref9, ref10}. In this study, the target grasping force is planned based on Coulomb’s friction law. The tangential force \( F_T \) between the gripper and the object is measured, and the coefficient of friction between the object and the fingers is \( \mu \). According to Coulomb's friction law, the minimum normal force required to prevent the object from sliding is \( F(t) = F_T(t)/\mu \). Based on this, the planned target grasping force is:
\begin{equation}
    F_d(t) = \max\left( \frac{1.2 F_T(t)}{\mu}, 1 \, \mathrm{N} \right) \label{eq33}
\end{equation}
Here an initial loading force of \( \text{1} \, \mathrm{N} \) is set, and a \( \text{20}\% \) tolerance is added. 

The force tracking component of the grasping algorithm uses the method proposed in this paper to control the actual normal force \( F(t) \) to follow the target force \( F_d(t) \).

\begin{figure}[t]
    \centering
    \includegraphics[width=8.8cm]{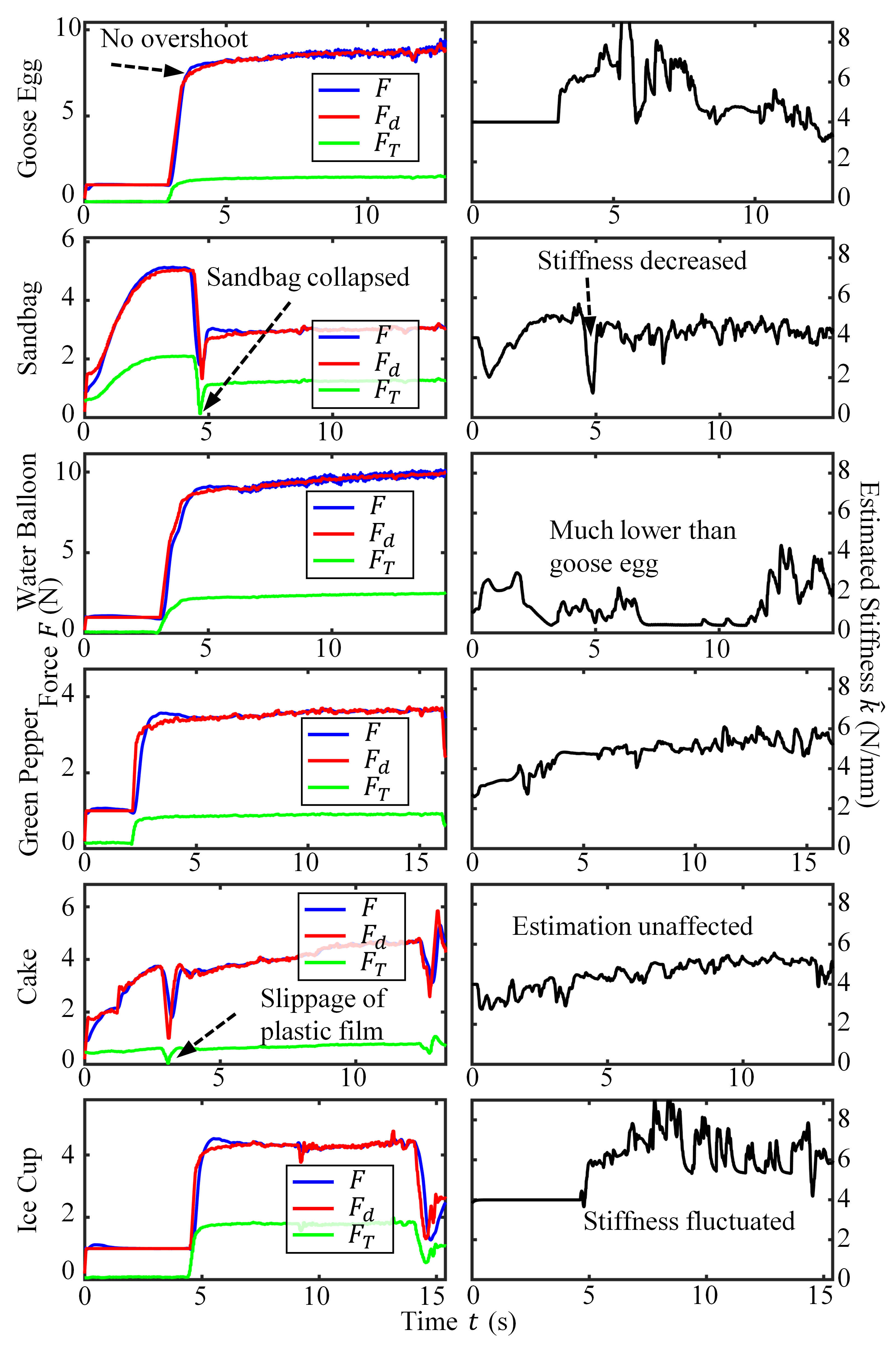}
    \caption{{Grasping experimental results. $F$ is the grasping force, $F_d$ is the target force, and $F_T$ is the tangential force.}}
    \label{fig:grasping_process}
\end{figure}

The grasping experimental results are shown in Fig.~\ref{fig:grasping_process}. The changes in actual tangential force \( F_T \), target normal force \( F_d \), actual normal force \( F \), and stiffness estimate \( \hat{k} \) during the grasping of different objects are recorded. When grasping the goose egg, the actual grasping force closely matches the target force, with no significant overshoot or oscillation, thereby preventing damage to the goose egg. In contrast, when grasping the water balloon, the stiffness estimate is significantly lower than that of the goose egg, which aligns with the soft physical properties of the water balloon. The lower stiffness means the gripper needs to close faster to match the actual normal force with the target force. The small generalized stiffness estimate increases the PI controller parameters, allowing the actual grasping force to quickly match the target force. Because our method adjusts the force tracking controller parameters based on the generalized stiffness estimate, it ensures good force tracking performance for both the goose egg and the water balloon, which have vastly different characteristics.

When grasping the sandbag, at around \( t = \text{4} \, \text{s} \), the robotic arm lifts the sandbag, and the sand particles lose their support from the table and flow toward the low points, causing the sandbag's shape to change. This leads to brief poor contact and relative slippage between the gripper and the sandbag. During this period, the stiffness estimate drops sharply, which corresponds with the small stiffness characteristic when the sand particles are flowing. Once the sandbag is fully lifted and stabilized, the gripper compresses the sandbag again, and the stiffness estimate increases. A similar behavior is observed with the cake. When the robotic arm lifts the cake, the tangential force measurement briefly approaches zero. But this is not due to the sand particles' flow but rather the temporary slippage of the cake’s plastic film surface. This does not affect the cake's stiffness, and thus the stiffness estimate remains stable. When grasping the ice cup, the gripper deforms the plastic shell and contacts the ice cubes, leading to complex stiffness changes due to the internal sliding of the ice cubes. The ice cup exhibits behavior that is quite different from that of an ideal variable-stiffness object and does not conform to the Burgers model or other physical models. Despite these special object behaviors and the complex physical processes they undergo, the stiffness estimator provides reasonable and accurate stiffness estimates, ensuring good force tracking performance, demonstrating the robustness of the method.

\begin{figure}[t]
    \centering
    \includegraphics[width=8.8cm]{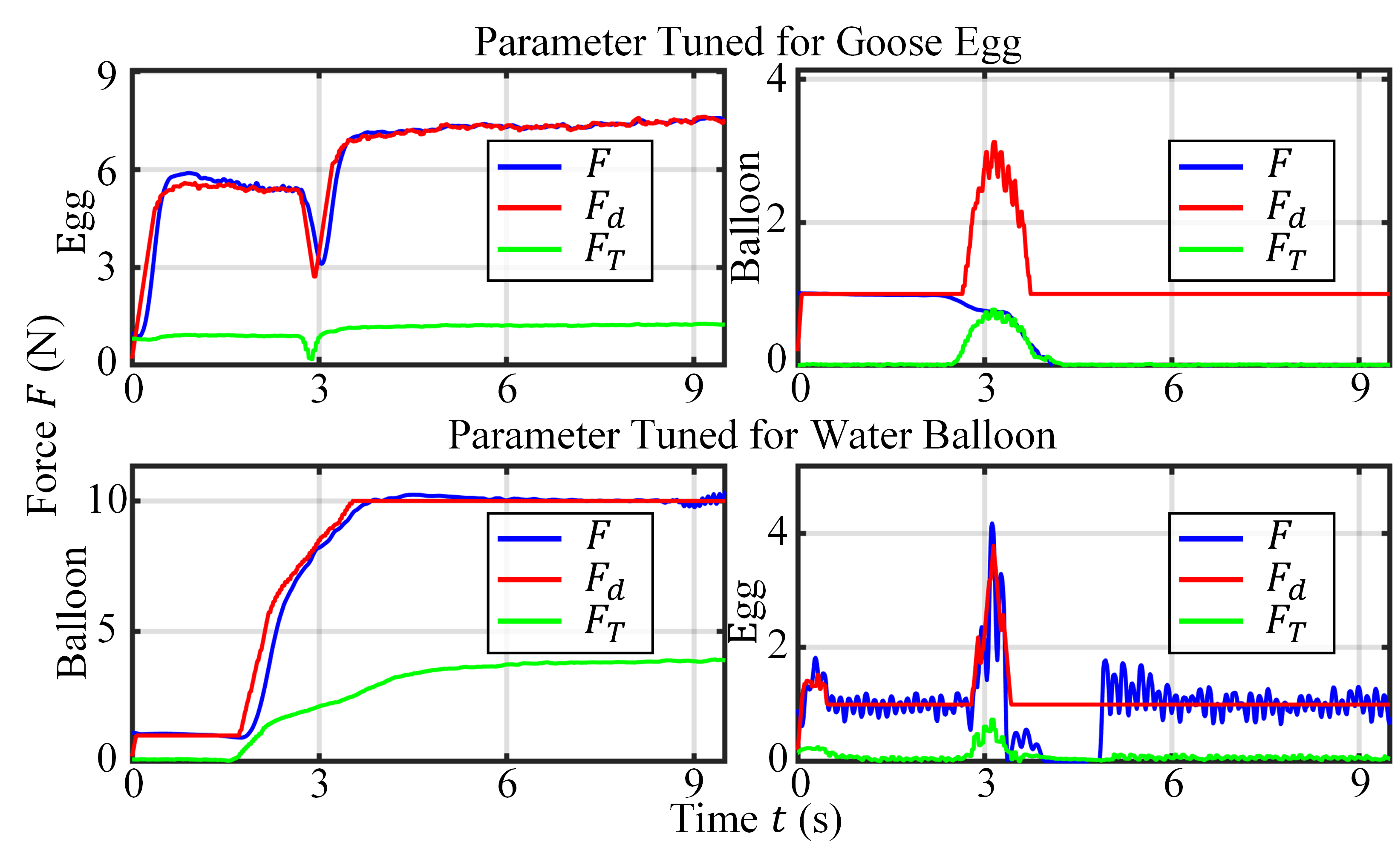}
    \caption{{Force tracking results with fixed parameters. $F$ is the grasping force, $F_d$ is the target force, and $F_T$ is the tangential force.}}
    \label{fig:fixed_parameters_results}
\end{figure}

Next, we perform an ablation experiment by removing the generalized stiffness estimator from our method and using manually preset parameters instead. {The manual adjustment process can be performed easily using conventional PI tuning methods because, as shown in} Equation~\eqref{eq:control_law}, {manually tunning $\hat{k}$ is essentially tunning the PI parameters of a standard PI controller.} We then conduct grasping experiments on several objects to verify the significance of online stiffness estimation in the grasping process. Initially, we manually adjust the force control parameters for the goose egg and find that the best force tracking performance occurs when the stiffness estimate is set to \( \hat{k} = \text{5} \, \text{N/mm} \). Grasping experiments are conducted on both the goose egg and water balloon, with the force planning algorithm and other conditions remaining the same as before. The experimental results are shown in Fig.~\ref{fig:fixed_parameters_results}. While force tracking works well for the goose egg, the performance for the water balloon is inadequate. Despite the target force increasing quickly when the robotic arm lifts the water balloon, the force tracking component fails to quickly adjust the actual grasping force to match the target, causing the object to slip.

Next, we adjust the parameters for the water balloon and set \( \hat{k} = \text{0.8} \, \text{N/mm} \) for the best force tracking performance. Grasping experiments on the water balloon and goose egg are conducted, and the results are shown in Fig.~\ref{fig:fixed_parameters_results}. While the force tracking works well for the water balloon, significant oscillations occur when grasping the goose egg. The low points of the oscillation lead to insufficient grasping force, causing the object to slip, while the high points lead to excessive grasping force, risking damage to the object and the gripper.

{These results are consistent with the theoretical analysis derived from} Equation~\eqref{eq:control_law}, {which indicates that an excessively large $\hat{k}$ results in smaller effective PI coefficients, leading to slow convergence. Conversely, a small $\hat{k}$ can cause the PI parameters to become too large, resulting in overshooting or oscillation.} In summary, the generalized stiffness estimation plays a crucial role in our force tracking algorithm and the overall grasping task. Only with accurate generalized stiffness estimates can good force tracking performance be guaranteed, enabling fast grasping and ensuring that the object does not slip or get damaged. This further demonstrates that our force tracking algorithm, which automatically adjusts the parameters based on generalized stiffness, is significantly superior to traditional methods without adaptive parameter tuning.

\section{Conclusion}
\label{sec:chapter6}
This paper proposes an adaptive grasping force tracking strategy based on
generalized stiffness, addressing the challenges of grasping force control in unstructured environments, where object models are unknown and nonlinear time-varying behaviors exist. By introducing the concept of generalized stiffness, this work breaks the limitation of traditional stiffness, which is defined only for ideal elastic objects, and enables adaptation to objects with complex mechanical behaviors. The LSTM-based generalized stiffness estimator designed in this study can accurately estimate stiffness online, with an average error of approximately one-thousandth of that from the commonly used least squares estimator, and can be used to adaptively adjust the parameters of the force tracking controller, achieving high precision and short probing time for grasping force tracking.

Force tracking experiments on \( \text{14} \) objects show that the proposed method performs well on elastic bodies, viscoelastic bodies, plastic objects, and variable-stiffness objects, with an average asymptotic error of \( \text{0.16} \, \text{N} \), about \( \text{1/3} \) of that achieved by the existing method MiFREN \cite{ref17}, and a probing time of only \( \text{3.34} \, \text{s} \), less than \( \text{1/11} \) of MiFREN's probing time. Our force tracking algorithm has also shown excellent application results in grasping tasks. For six common non-ideal objects, which are difficult to control in grasping force, successful grasping was achieved through superior force tracking performance. {These experiments demonstrate successful generalization of our neural network from training data to real-world applications.} Ablation experiments removing the generalized stiffness estimator demonstrated the irreplaceable role of our stiffness estimator.

Future research will focus on optimizing the neural network structure to reduce computational costs and further expanding the experimental objects and application scenarios. With these improvements, the proposed adaptive grasping force tracking method is expected to play an even greater role in robotic autonomous grasping tasks.


\bibliographystyle{IEEEtran}
\bibliography{references}

\newpage
 
\vspace{11pt}
\newcommand{\bioimage}[1]{%
  \adjustbox{valign=t, width=2.54cm, height=3.26cm, keepaspectratio}{%
    \includegraphics{#1}%
  }%
}
\begin{IEEEbiography}[{\bioimage{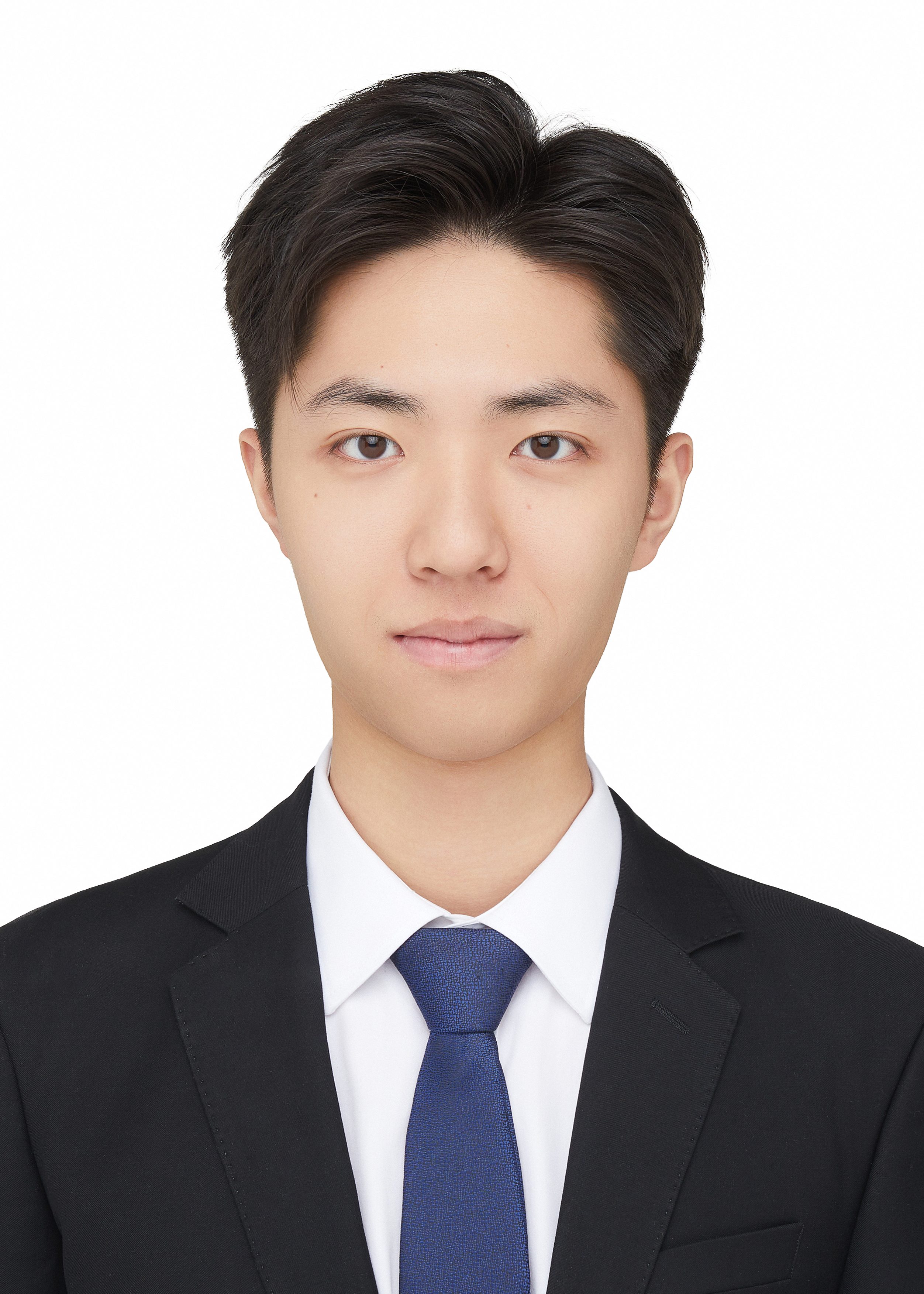}}]{Ziyang Cheng}
was born in Wenzhou, Zhejiang, China, in 2003. He is currently pursuing the B.S. degree in mathematics and physics and the B.E. degree in mechanical engineering at Tsinghua University, Beijing, China.

His research interests include robotics and machine learning.
\end{IEEEbiography}

\begin{IEEEbiography}[{\bioimage{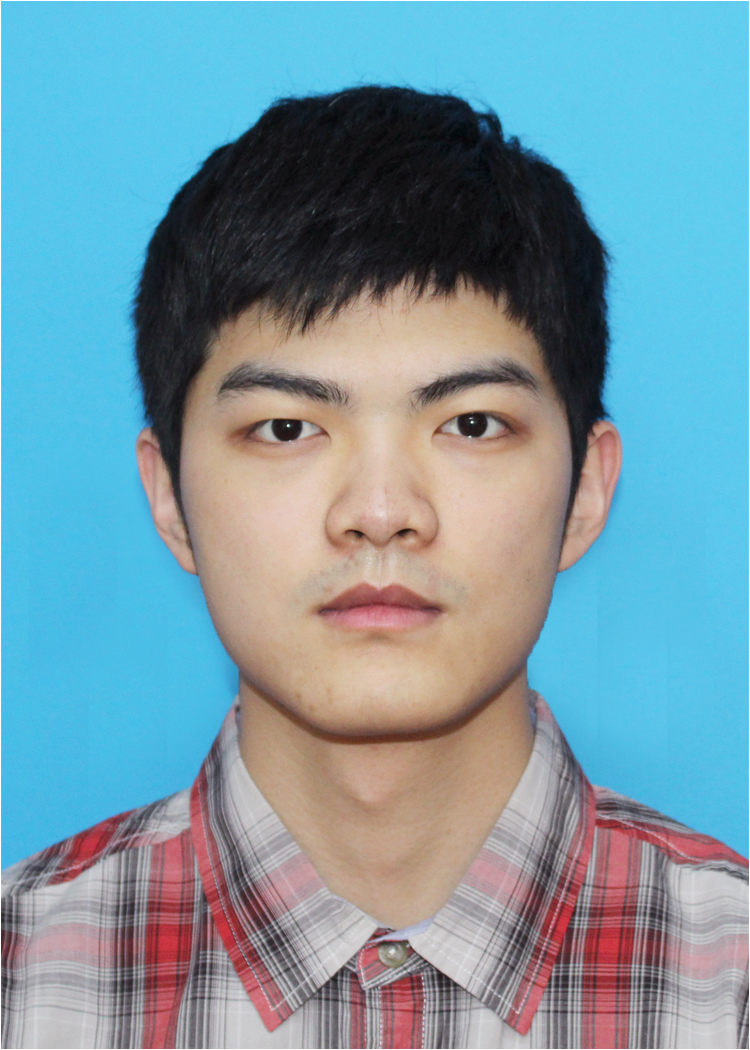}}]{Xiangyu Tian}
received the B.E. degree in mechanical engineering in 2020 from Tsinghua University, Beijing, China, where he is currently working toward the Ph.D. degree in mechanical engineering.

His research interests include robot control and visual servoing.
\end{IEEEbiography}

\begin{IEEEbiography}[{\bioimage{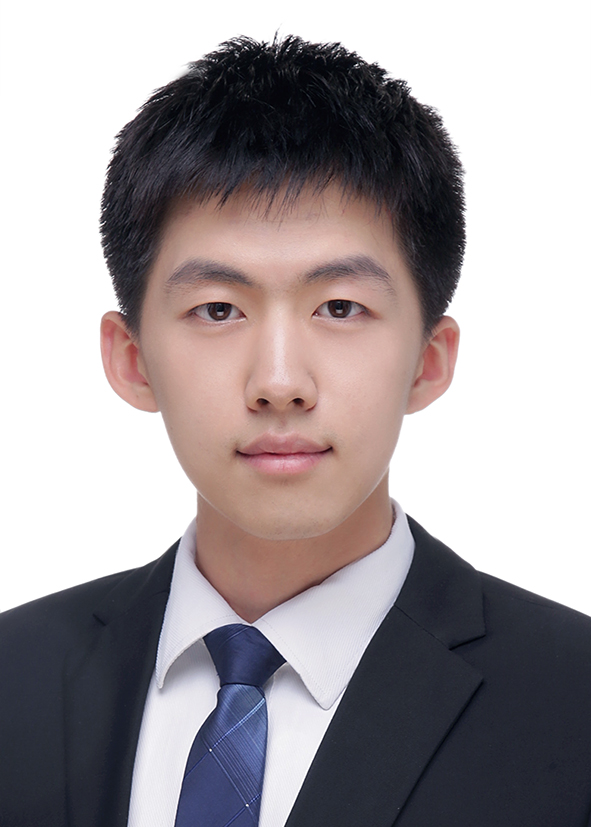}}]{Ruomin Sui}
received the B.E. degree and the Ph.D. degree from Tsinghua University, Beijing, China, in 2019 and 2024, all in mechanical engineering.

His research interests include human tactile physiology and tactile sensing techniques, particularly incipient slip detection and its application in robotic grip control.
\end{IEEEbiography}

\begin{IEEEbiography}[{\bioimage{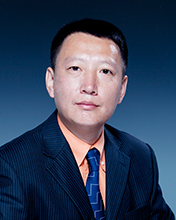}}]{Tiemin Li}
received the B.S. degree from the Qiqihar Institute of Light Industry, Heilongjiang, China, in 1991, the M.S. degree from the Harbin Institute of Technology, Heilongjiang, China, in 1996, and the Ph.D. degree from Tsinghua University, Beijing, China, in 2000, all in mechanical engineering.

He is currently an Associate Professor with the Department of Mechanical Engineering, Tsinghua University. His current research interests include advanced manufacturing technology, robotics, and parallel kinematic machines.
\end{IEEEbiography}

\begin{IEEEbiography}[{\bioimage{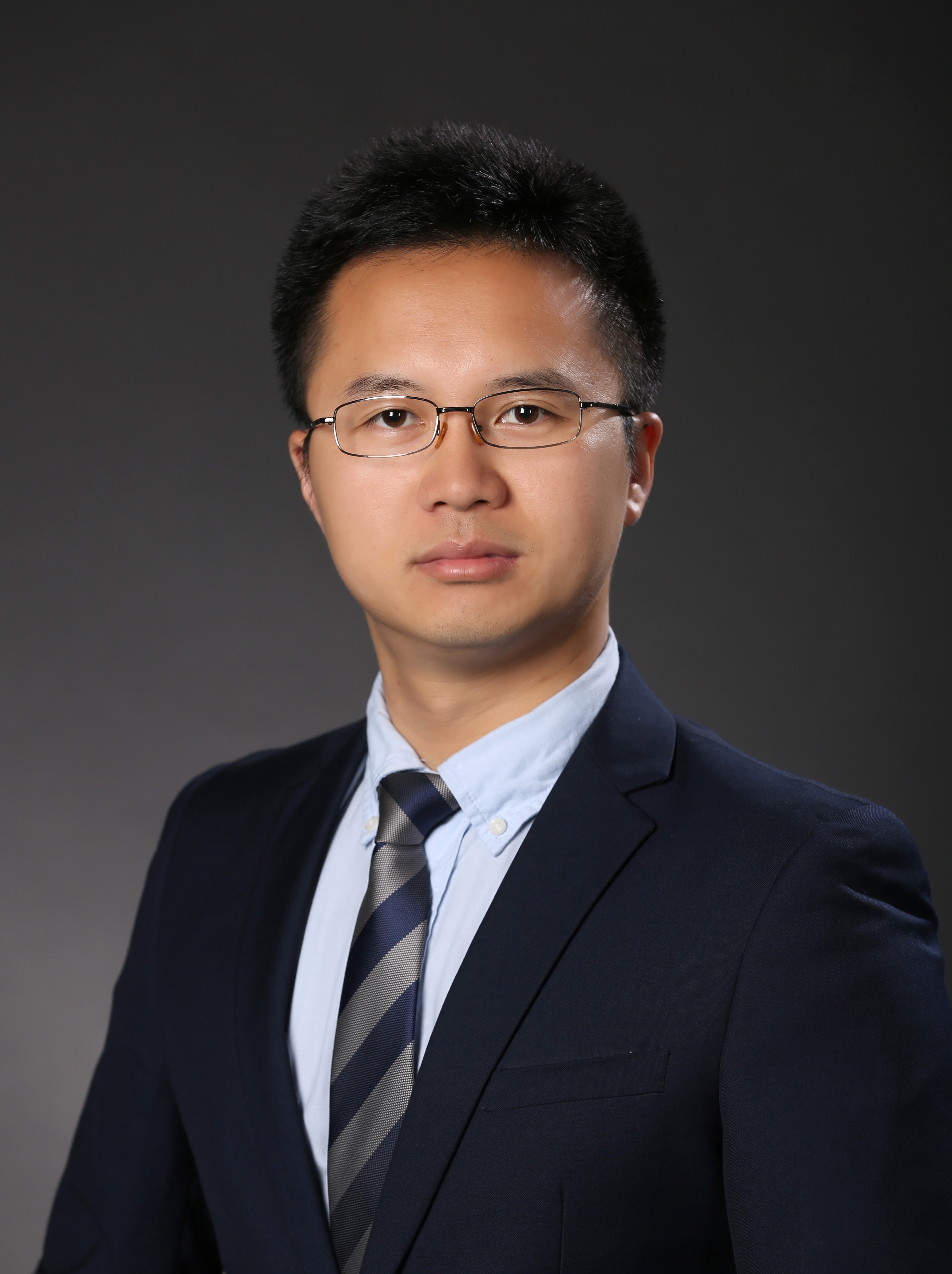}}]{Yao Jiang}
(Member, IEEE) received the B.S. degree from the Nanjing University of Science and Technology, Nanjing, China, in 2011, and the Ph.D. degree from Tsinghua University, Beijing, China, in 2016, all in mechanical engineering.

From 2016 to 2017, he was a Postdoctoral Researcher with the Department of Precision Instrument, Tsinghua University. From 2017 to 2018, he was a Postdoctoral Researcher with the Harvard School of Engineering and Applied Sciences, Cambridge, MA, USA. He is currently an Associate Professor with the Department of Mechanical Engineering, Tsinghua University. His current research interests include robotic autonomous manipulation, tactile sensors, computer vision metrics, and mobile robots.
\end{IEEEbiography}

\vfill

\end{document}